\documentclass{article}

\usepackage[final, nonatbib]{icbinb2021}

\usepackage[utf8]{inputenc} 
\usepackage[T1]{fontenc}    
\usepackage{hyperref}       
\usepackage{url}            
\usepackage{booktabs}       
\usepackage{amsfonts}       
\usepackage{nicefrac}       
\usepackage{microtype}      
\usepackage{xcolor}         

\usepackage{amsmath}
\usepackage{subcaption}
\usepackage{graphicx}
\usepackage{float}
\usepackage{wrapfig, lipsum,booktabs}

\usepackage{enumitem}

\title{Understanding Out-of-distribution:A Perspective of Data Dynamics}

\author{%
  Dyah Adila\thanks{Corresponding author} \\
  Department of Computer Sciences\\
  University of Wisconsin-Madison\\
  \texttt{adila@wisc.edu} \\
   \And
   Dongyeop Kang \\
    Department of Computer Science and Engineering\\
    University of Minnesota\\
    \texttt{dongyeop@umn.edu}\\
}

\begin{document}

\maketitle

\begin{abstract}
  Despite machine learning models' success in Natural Language Processing (NLP) tasks, predictions from these models frequently fail on out-of-distribution (OOD) samples. Prior works have focused on developing state-of-the-art methods for detecting OOD. The fundamental question of how OOD samples differ from in-distribution samples remains unanswered. This paper explores how data dynamics in training models can be used to understand the fundamental differences between OOD and in-distribution samples in extensive detail. We found that syntactic characteristics of the data samples that the model consistently predicts incorrectly in both OOD and in-distribution cases directly contradict each other. In addition, we observed preliminary evidence supporting the hypothesis that models are more likely to latch on trivial syntactic heuristics (e.g., overlap of words between two sentences) when making predictions on OOD samples. We hope our preliminary study accelerates the data-centric analysis on various machine learning phenomena.
\end{abstract}

\section{Introduction}

Detecting out-of-distribution (OOD) has become one of the key bottlenecks in building reliable open-world systems \cite{bendale2015towards}, leading to new SOTA approaches meant to mitigate this problem \cite{chen2021robustifying, 10.5555/3295222.3295387, lee2018simple, tayal2020model,liu2020energy, mohseni2020self}.
Previous works have explored the discrepancy between OOD and in-distribution test performance from the model and algorithm perspective.
Le Lan \& Dinh (2020) \cite{lan2020perfect} observed the limitation of density estimation for anomaly detection. Zhang et al. (2021) \cite{zhang2021understanding} suggested that the cause of OOD failure is the combination of the model architecture
and maximum likelihood objective. Choi et al. (2018) \cite{Choi2018WAICBW}, Just \& Ghosal (2019) \cite{just2019deep}, Fetaya et al. (2020) \cite{fetaya2019understanding}, Kirichenko et al. (2020) \cite{kirichenko2020normalizing}, Zhang et al. (2020) \cite{10.1007/978-3-030-58580-8_7}, and Wang et al. (2020) \cite{wang2020analysis} concluded that significant mismatch between the real and estimated distribution gives rise to the performance discrepancy. Nevertheless, what are the fundamental differences between in-distribution and OOD data samples which cause models to be especially brittle to the latter?

Our approach to studying this difference is inspired by emerging studies on machine learning models adopting \textit{shallow heuristics} (i.e., irrelevant statistical patterns found in the majority of training examples), instead of learning the underlying generalizations that they are intended to capture \cite{mccoy-etal-2019-right, wang2017visual, agrawal-etal-2016-analyzing, 10.1007/978-3-030-01270-0_28}. For example, Beery et al. (2018) \cite{10.1007/978-3-030-01270-0_28} demonstrated a network that accurately recognizes cows in a typical context (e.g., pasture) consistently misclassifies cows in a non-typical context (e.g., beach). Similar heuristics also arise in visual question answering
systems \cite{agrawal-etal-2016-analyzing} and researchers proposed graph generative modeling schemes \cite{jiang2021x} (inspired by graph convolutional networks \cite{tayal2020regularized}) to handle the problem implicitly. In this paper, we study this problem within the Natural Language Inference (NLI): the task of determining whether a premise sentence entails (i.e., implies the truth of) a hypothesis sentence \cite{condoravdi-etal-2003-entailment, 10.1007/11736790_9, bowman-etal-2015-large}. McCoy et al. (2019) \cite{mccoy-etal-2019-right} exhaustively characterized shallow heuristics that commonly appears in benchmark NLI datasets, which we will refer to as \textit{syntactic heuristics}.

To this end, we study the difference between in-distribution and OOD samples which gives rise to models brittleness through two perspectives: (i) \textit{training dynamics} of the model, and (ii) \textit{syntactic heuristics} of the data samples. Specifically, we perform two types of data-dynamic analyses: We first examine the difference between OOD and in-distribution samples distributions in the data cartography space \cite{swayamdipta-etal-2020-dataset} at each training epoch. Cartography space allows us to distinguish which samples are harder to learn (i.e., the model often misclassifies them) based on training dynamics measures. Second, we mark each data sample by their syntactic heuristic, which enables us to identify what shallow characteristics tend to be harder to learn. For instance, we hypothesize whether more word overlap between a premise and a hypothesis text influences model's ability to infer their label correctly.

Our analyses suggested that the syntactic heuristic that the model deems hard-to-learn during training directly contradicts the characteristic of OOD samples that are hard-to-learn during inference. We also found preliminary evidence suggesting the model's tendency to make inferences based on trivial syntactic heuristic is higher in the OOD case. We hope that this study will drive more effort to better understand the difference between in-distribution and OOD samples and, consequently, develop more informed and data-centric OOD detection and generalization methods.

\section{Our Hypotheses}
This section presents our two initial hypotheses to characterize OOD data samples.
Motivated by previous works \cite{hendrycks17baseline,hendrycks-etal-2020-pretrained}, we perform a comparative analysis between OOD and in-distribution samples over \textit{dynamic information of model training} (e.g., confidence, prediction variance across epoch) per set of train-test epoch. Moreover, to interpret the quantitative difference of these measures, we mark data samples based on their \textit{syntactic heuristics}. We hypothesize that the difference between OOD and in-distribution samples can be characterized in the combination of multiple training dynamics (\S\ref{sec:h1}) and sample heuristics (\S\ref{sec:hypothesis-heuristic}).

\subsection{\textsc{H1}: Are OOD and in-distribution samples different with their training dynamics?}\label{sec:h1}

Data cartography \cite{swayamdipta-etal-2020-dataset} is a tool to group data samples into three regions: easy-to-learn, hard-to-learn, and ambiguous (see Figure \ref{fig:result-ex1}), enabled by following two training dynamics as axes: 
\begin{enumerate}[leftmargin=*]
    \item \textbf{Confidence}: the mean model probability of the true label ($y_i^*$) across epochs (equation \ref{eq:confidence}). The term $p_{\theta}^{(e)}(y_i^* | x_i)$ denotes the model's probability with parameters $\theta$ at the end of $e^{th}$ epoch.
    
    \item \textbf{Variability}: the spread of $p_{\theta}^{(e)}(y_i^* | x_i)$ across epochs, and is defined by equation \ref{eq:variability}. 
\end{enumerate}

\begin{minipage}{.4\linewidth}
\begin{equation}
\label{eq:confidence}
    \hat{\mu}_i = \frac{1}{E}\sum_{e=1}^{E}p_{\theta}^{(e)}(y_i^* | x_i)
\end{equation}
\end{minipage}%
\medskip
\begin{minipage}{.5\linewidth}
\begin{equation}
\label{eq:variability}
    \hat{\sigma_i}=\sqrt{\frac{\sum_{e=1}^{E}(p_{\theta}^{(e)}(y_i^* | x_i)-\hat{\mu}_i)^2}{E}}
\end{equation}
\end{minipage}

Intuitively, higher variability implies a high range of probability outputted by the model for the same sample. As seen in Figure \ref{fig:result-ex1}, the rightmost region (i.e., high variability) is the ambiguous region. The easy-to-learn region is characterized by high confidence and low variability (i.e., correct prediction with high assigned probability across epochs), and hard-to-learn region samples have low confidence and low variability (i.e., incorrect prediction across epochs).

At each epoch $E$, we record the following four measurements for all train samples $i \in N_{train}$ and test samples $j \in N_{test}$ ($ N_{train}$ and $ N_{test}$ denote the size of train and test sets respectively).  As $\theta$ is constant for all samples at each epoch, for conciseness sake, we abbreviate $p_{\theta}^{(e)}(y_i^* | x_i)$ as $p_{i}^{(e)}$.

\begin{minipage}{.3\linewidth}
\begin{equation}
\label{eq:confidence_train}
    \hat{\mu}_{i}^{E} = \frac{1}{E}\sum_{e=1}^{E}p_{i}^{(e)}
\end{equation}
\end{minipage}%
\medskip
\begin{minipage}{.4\linewidth}
\begin{equation}
\label{eq:variability_train}
    \hat{\sigma}_{i}^{E}=\sqrt{\frac{\sum_{e=1}^{E}(p_{i}^{(e)}-\hat{\mu}_{i}^{E})^2}{E}}
\end{equation}
\end{minipage}
\begin{minipage}{.2\linewidth}
\begin{equation*}
    \forall \quad i \in N_{train}
\end{equation*}
\end{minipage}%

\begin{minipage}{.3\linewidth}
\begin{equation}
\label{eq:confidence_test}
    \hat{\mu}_j^{E} = \frac{1}{E}\sum_{e=1}^{E}p_{j}^{(e)}
\end{equation}
\end{minipage}%
\medskip
\begin{minipage}{.4\linewidth}
\begin{equation}
\label{eq:variability_test}
    \hat{\sigma_j}^{E}=\sqrt{\frac{\sum_{e=1}^{E}(p_{j}^{(e)}-\hat{\mu}_j^E)^2}{E}}
\end{equation}
\end{minipage}
\begin{minipage}{.2\linewidth}
\begin{equation*}
    \forall \quad j \in N_{test}
\end{equation*}
\end{minipage}%

In each epoch $E$, we then plot all train samples based on their respective ($\hat{\mu}_i^{E}$, $ \hat{\sigma_i}^{E}$) resulting in train cartography map $C_{train}^{E}$ (e.g., figure \ref{fig:result_ex1-train}). Similarly, we also plot all test samples based on their ($\hat{\mu}_j^{E}$, $ \hat{\sigma_j}^{E}$) resulting in test cartography map $C_{test}^{E}$. (e.g., figure \ref{fig:result_ex1-test})


As models' performance in inferring in-distribution samples far exceeds that of the OOD case, we hypothesize that most OOD samples will stay in the hard-to-learn regions in $C_{test}^{E}$ for all $E$ . In contrast, most in-distribution samples will eventually move to the easy-to-learn region in $C_{test}^{E}$ for larger $E$, which results in a \textit{noticeable difference between the two distributions in the cartography space.} We hope to interpret this difference using syntactic heuristic (\S\ref{sec:hypothesis-heuristic}).

\subsection{\textsc{H2}: Do models tend to adopt syntactic attributes of OOD samples more readily?}

\label{sec:hypothesis-heuristic}

Past works have shown that machine learning models often adopt `shortcut' data characteristics instead of the generalization that humans would learn and use to perform the same task \cite{mccoy-etal-2019-right, wang2017visual, agrawal-etal-2016-analyzing, 10.1007/978-3-030-01270-0_28}. More specifically, a model trained on NLI might assign a label of contradiction to any input containing the word ``not'' \cite{naik-etal-2018-stress,sanchez-etal-2018-behavior}, as this heuristic often applies in standard NLI training sets. 
As OOD samples come from a different distribution, they will have less common generalizable characteristics with the training set than the in-distribution samples. Thus, we hypothesize that \textit{models will rely even more upon the shortcut characteristics (i.e., heuristics) when making inferences on OOD samples.}

\begin{figure}
\vspace{-3mm}
\small
  \begin{subfigure}[t]{.5\textwidth}
    \centering
    \includegraphics[width=\linewidth,trim={0 0 0 1.2cm},clip]{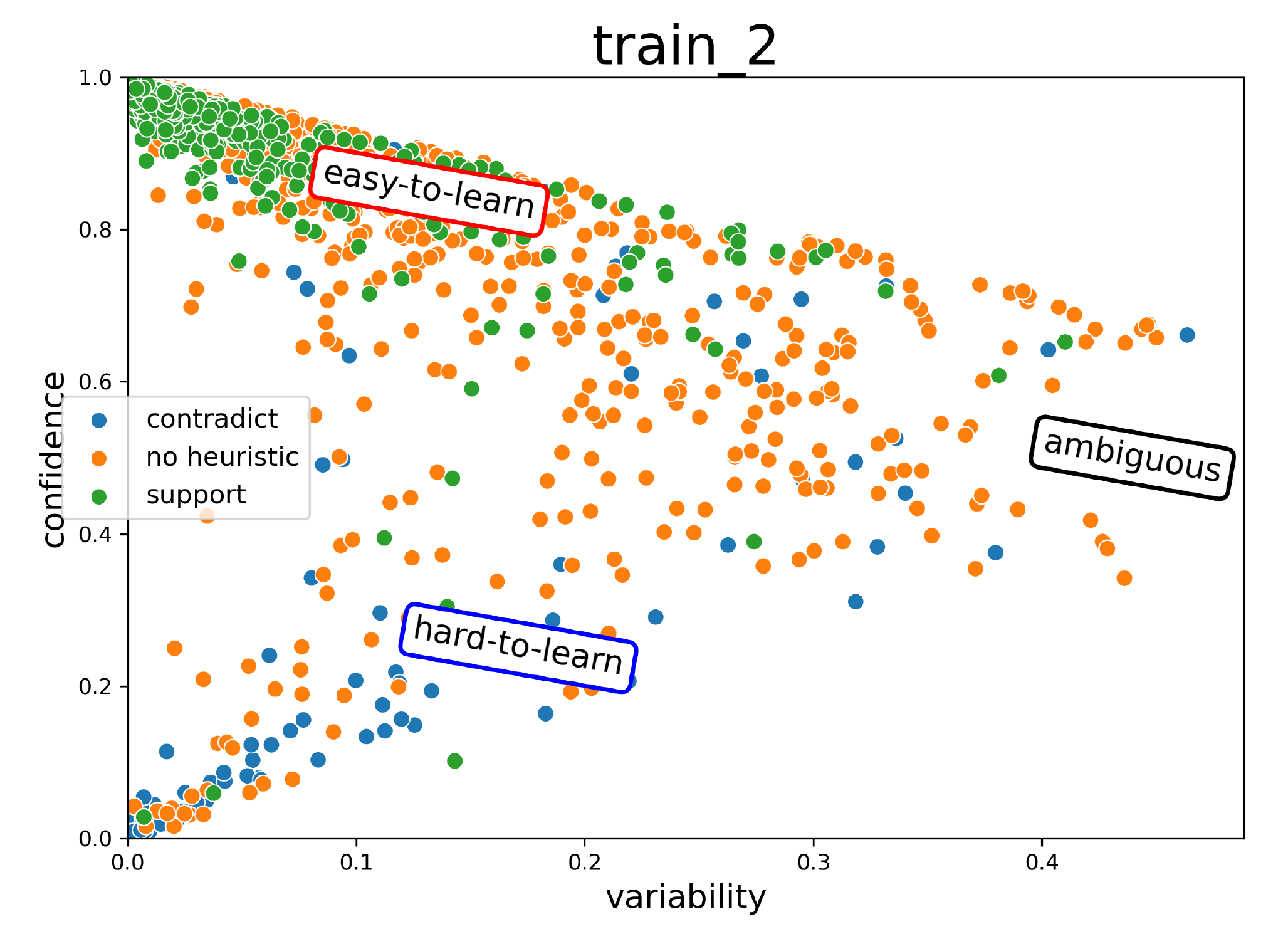}
    \caption{\textbf{Training} cartography map at \textbf{$E=2$} ($C_{train}^2$)}
    \label{fig:result_ex1-train}
  \end{subfigure}
  \hfill
  \begin{subfigure}[t]{.5\textwidth}
    \centering
    \includegraphics[width=\linewidth,trim={0 0 0 1.2cm},clip]{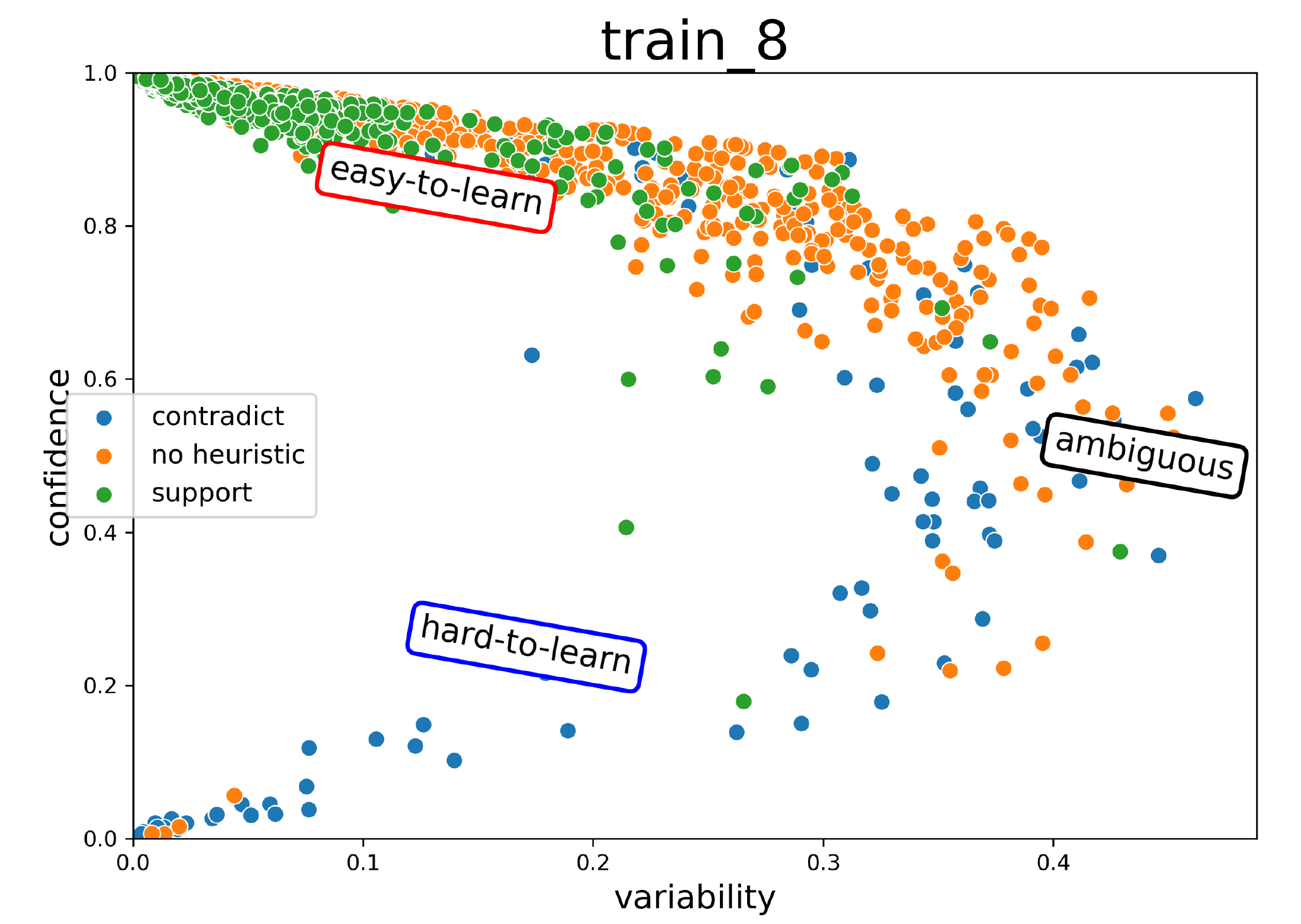}
    \caption{\textbf{Training} cartography map at \textbf{$E=8$} ($C_{train}^8$)}
  \end{subfigure}
  \medskip
  \begin{subfigure}[t]{.5\textwidth}
    \centering
    \includegraphics[width=\linewidth,trim={0 0 0 1.2cm},clip]{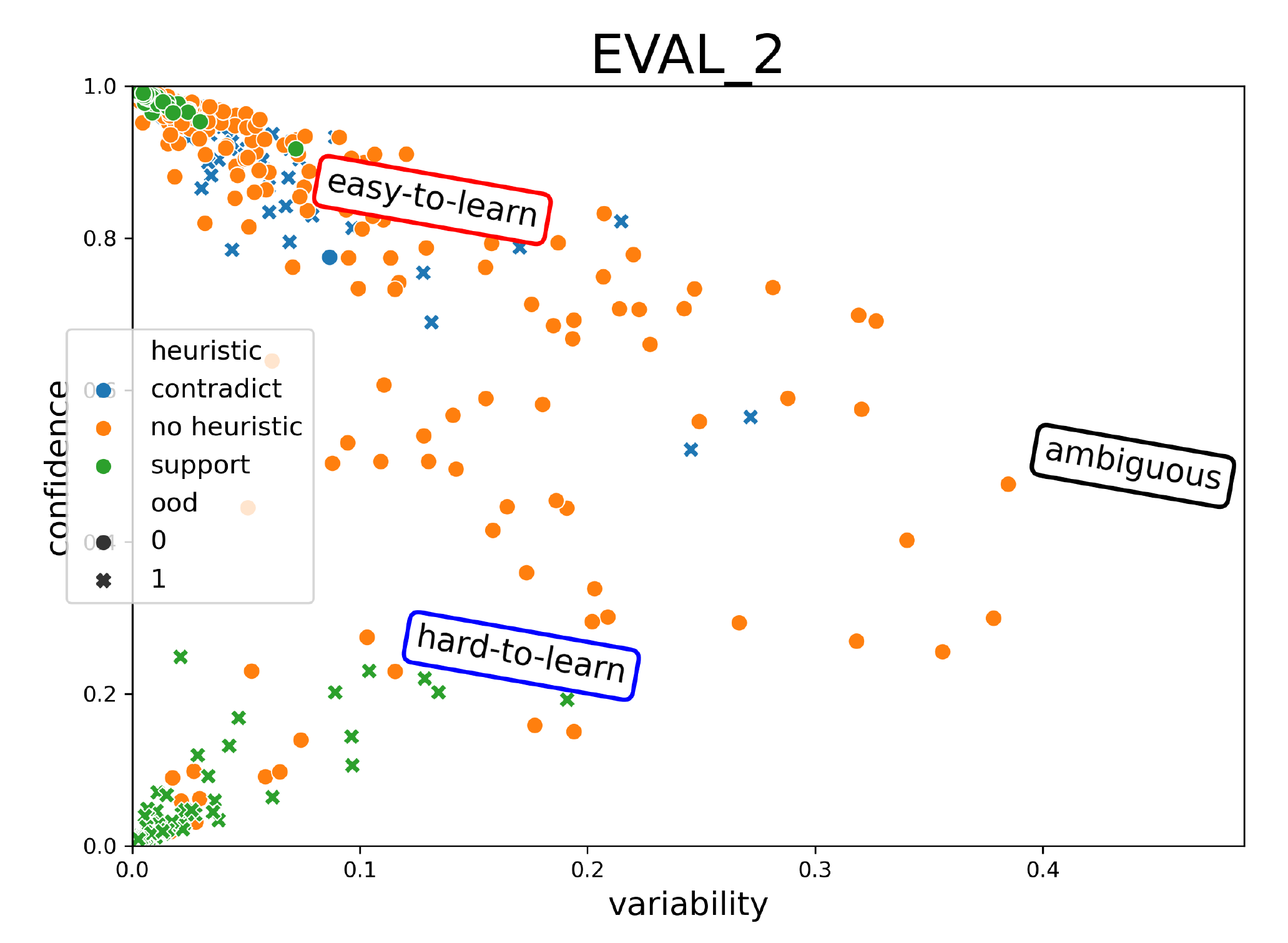}
    \caption{\textbf{Evaluation} cartography map at \textbf{$E=2$} ($C_{test}^2$)}
    \label{fig:result_ex1-test}
  \end{subfigure}
  \hfill
  \begin{subfigure}[t]{.5\textwidth}
    \centering
    \includegraphics[width=\linewidth,trim={0 0 0 1.2cm},clip]{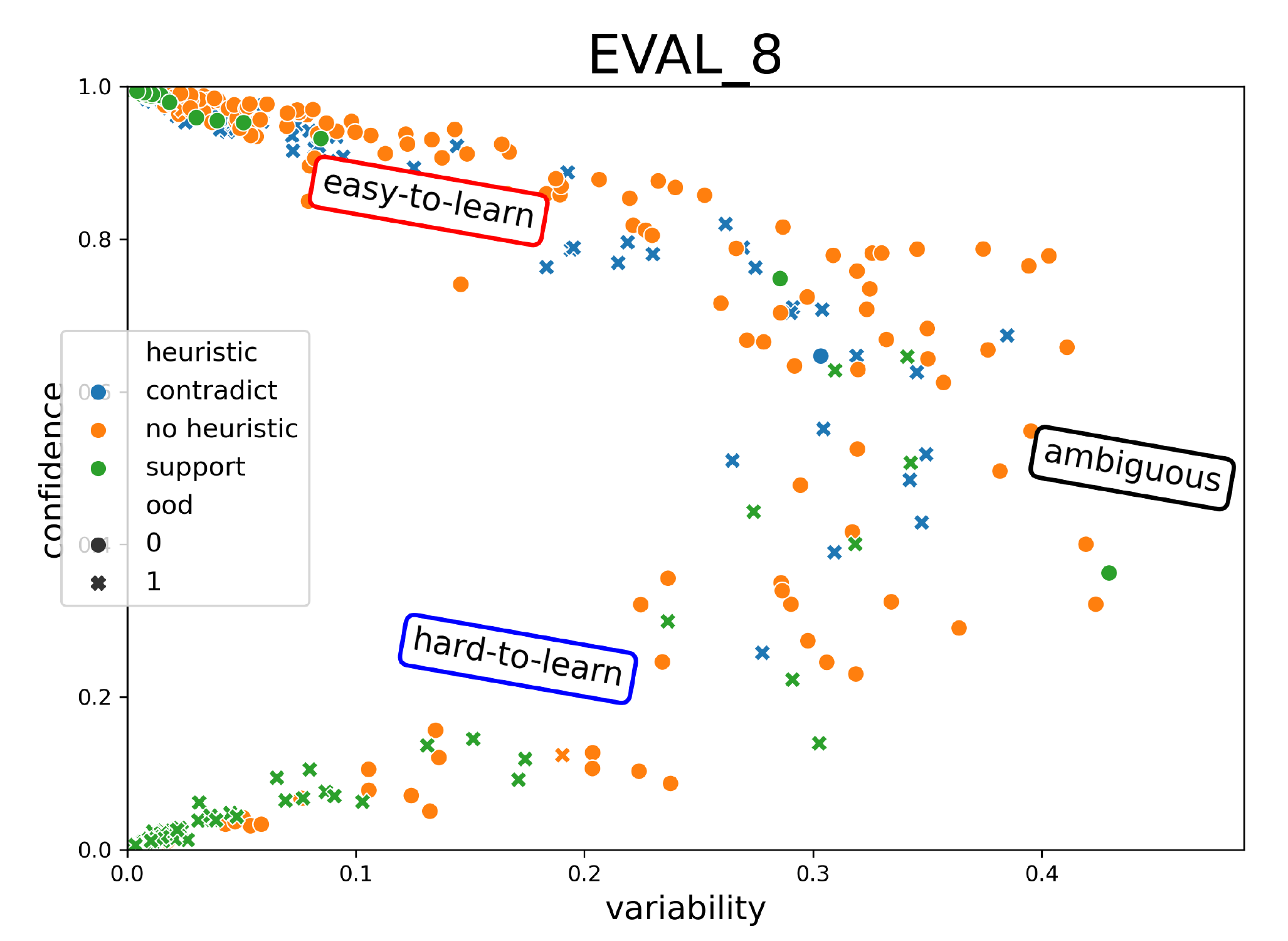}
    \caption{\textbf{Evaluation} cartography map at \textbf{$E=8$} ($C_{test}^8$)}
  \end{subfigure}
  \vspace{-3mm}
 \caption{Training and evaluation cartography maps. The different colors illustrate different sample heuristics and the different shapes in figures (c) and (d) illustrate OOD sample ('o' for in-distribution and 'x' for OOD).}
 \label{fig:result-ex1}
 \vspace{-.5cm}
\end{figure}

We focus on lexical overlap heuristic \cite{mccoy-etal-2019-right}, as it is one of the simplest and most common across NLI datasets. This heuristic assumes that a premise \textit{entails} all hypotheses constructed from words in the premise. Below are examples:
\\
\vspace{1mm}
\begin{tabular}{cccc}
\centering
\footnotesize
Premise     & Hypothesis     & Label & Type \\
\midrule
The judge was paid by the actor. & The actor paid the judge.  & Entailed   & Support  \\
The actor was paid by the judge.     & The actor paid the judge. & Not entailed & Contradict     \\
\vspace{1mm}
\end{tabular}
 \\
A sample can either: (i) supporting, (ii) contradicting (i.e., possessing the property of heuristic and labels \textit{entailed} or \textit{not entailed} respectively), or (iii) having no heuristic. We mark each sample's heuristics, as illustrated by the different circle colors of Figure \ref{fig:result-ex1}.

To quantify the degree of which each sample adopts lexical heuristic, we calculate the percentage of words in the hypothesis (s2) which overlap with premise (s1), which aligns with the lexical heuristic definition. More formally, we measure $m2 = \frac{\left | s1 \bigcap s2 \right |}{\left | s2 \right |}$. 

To measure the model's tendency to adapt the heuristic, we calculate three sets of correlations: (i) $\rho(m2, {\mu}_i^{E})$ (ii) $\rho(m2, {\mu}_{j(in-dist)}^{E})$ (iii) $\rho(m2, {\mu}_{j(OOD)}^{E})$, and observe the trends across epochs. We calculate these correlations for all samples and also for samples from each class independently. Intuitively, a high heuristic adoption is indicated by high absolute correlation values in the individual class samples (e.g., the model learns that high words overlap means the label is \textit{entailment}, vice versa).

\section{Experiments}

\begin{wraptable}{r}{0.5\textwidth}
\vspace{-4mm}
 \small
  \caption{Datasets combination in our experiments.}
  \label{tab:experiment-data}
  \centering
  \begin{tabular}{@{}lll@{}}
    \toprule
    Training & Eval (In-distribution) & Eval (OOD) \\
    \midrule
    MNLI (train)  & MNLI (dev matched) & WNLI (train)  \\
    MNLI (train)  & MNLI (dev matched) & RTE (dev)    \\
    RTE (train) & RTE (dev) & WNLI (train)     \\
    \bottomrule
  \end{tabular}
  \vspace{-4mm}
\end{wraptable}
\textbf{Setup:} In all experiments, we trained Roberta \cite{liu2019roberta} with batch size 20 and learning rate $1.1e^{-5}$, and initialized with the same random seed. 
All experiments were carried out using PyTorch \cite{pytorch} with one NVidia Tesla K20X GPU. Experiments were carried out using three datasets (MNLI \cite{williams-etal-2018-broad}, RTE \cite{wang2019glue}, WNLI \cite{levesque2012winograd}) which combinations are shown in table \ref{tab:experiment-data}. 

In our study, we define OOD samples as those with different textual genre and sentence structure from the training. For instance, WNLI were derived by taking sentences from fiction books. Each hypothesis-premise pair is a sentence and it's copy with the ambiguous pronouns replaced by each possible referent. While RTE and MNLI's hypothesis-premise pairs are different sentences. RTE are texts from Wikipedia and MNLI were crowdsourced from speech, fiction, and government report. Our extensive experiments results can be found in Appendix \ref{appendix:exp-1-mnli}, Appendix \ref{appendix:exp-1-rte}, and Appendix \ref{appendix:exp-2-entailment}. For conciseness sake, we only present results with MNLI as train and in-distribution data, and WNLI as OOD data.

\subsection{OOD vs in-distribution on training dynamics and syntactic heuristic}
\label{sec:result-ex1}

Result for this section are the cartography maps $C_{train}^2$, $C_{train}^8$, $C_{test}^2$, $C_{test}^8$ in Figure \ref{fig:result-ex1}. For visualization clarity, the points showed are a sampled fraction from the original set. We observe the following:
\begin{itemize}[leftmargin=*]
    \item \textbf{Observation:} A contradicting pattern between the trajectories of training and OOD samples. By comparing figures (a) and (b) ($C_{train}^2$ and $C_{train}^8$), we observed samples that \textit{contradict the heuristic} (blue circles) either stays in the hard-to-learn region or move towards the ambiguous region. On contrary, the figures (c) and (d) ($C_{test}^2$ and $C_{test}^8$) shows that OOD samples that \textit{support the heuristic} (green 'x's) stay in the hard-to-learn region.
    \\
    \textbf{Conjecture:} Samples heuristics (i.e., lexical overlap) that the model deems hard/easy to learn during training \textit{completely flips in the OOD case}.
    \item \textbf{Observation:} At the end of epoch 8,  we observe  way less color mix at the ambiguous region of $C_{test}^8$ (blue and green 'x's at the ambiguous region of map (d)). While in $C_{test}^8$ (map (b)), this mix is very apparent in the easy-to learn region (blue and green circles).
    \\
    \textbf{Conjecture:} a more generalizable knowledge is learned during training (indicated by the samples mix). Although, \textit{this generalizable knowledge is slower to be adopted by the model in inferring OOD samples}.
\end{itemize}

\subsection{OOD vs in-distribution on tendency to adopt syntactic heuristics}
\label{sec:result-ex2}
\begin{wrapfigure}{r}{0.4\textwidth}
\vspace{-3mm}
\small
  \begin{subfigure}[t]{.4\textwidth}
    \centering
    \includegraphics[width=\linewidth,trim={0 0 0 .5cm},clip]{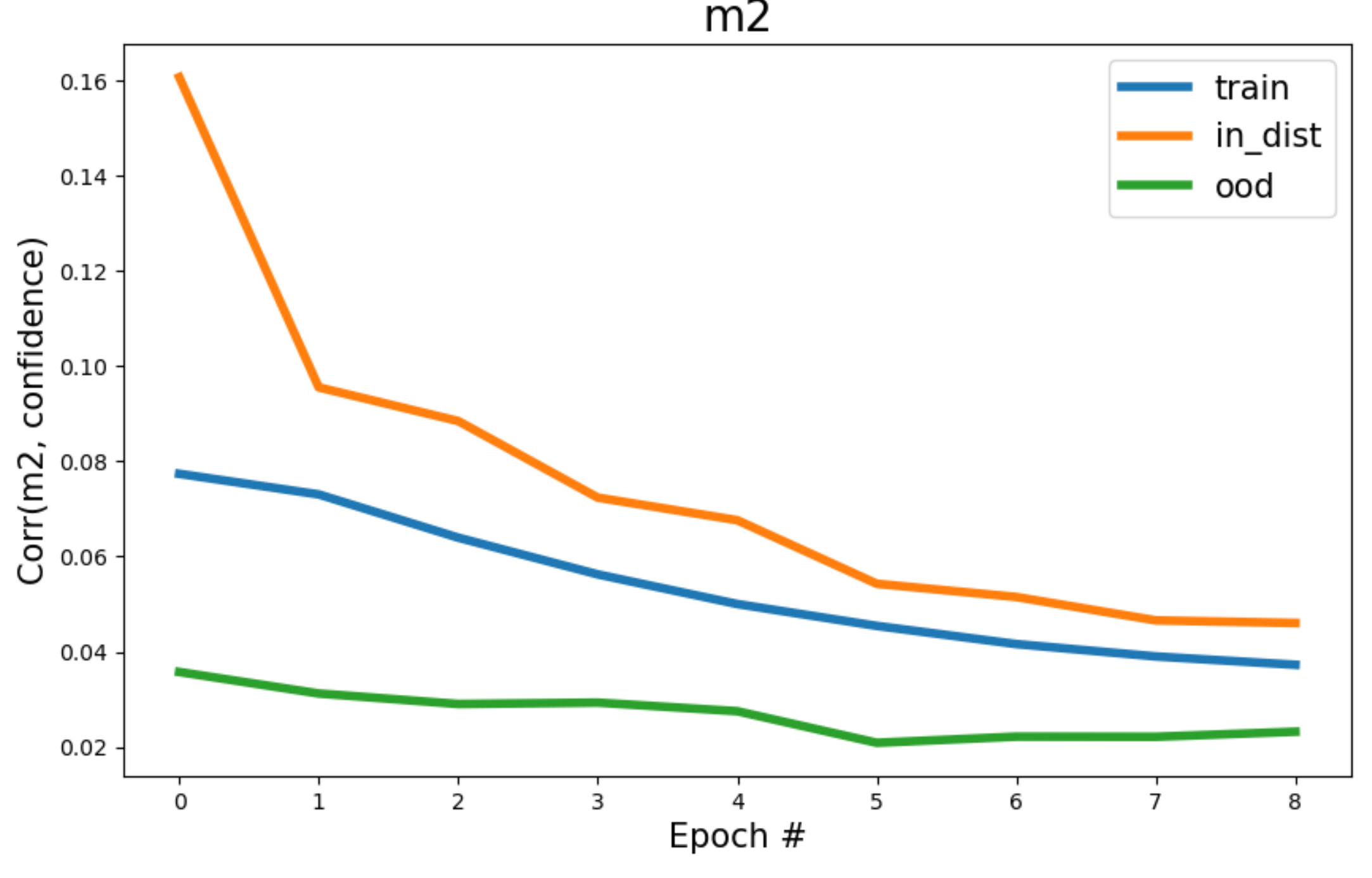}
    \caption{Correlation in \textbf{all} samples}
  \end{subfigure}
  \begin{subfigure}[t]{.4\textwidth}
    \centering
    \includegraphics[width=\linewidth,trim={0 0 0 1.1cm},clip]{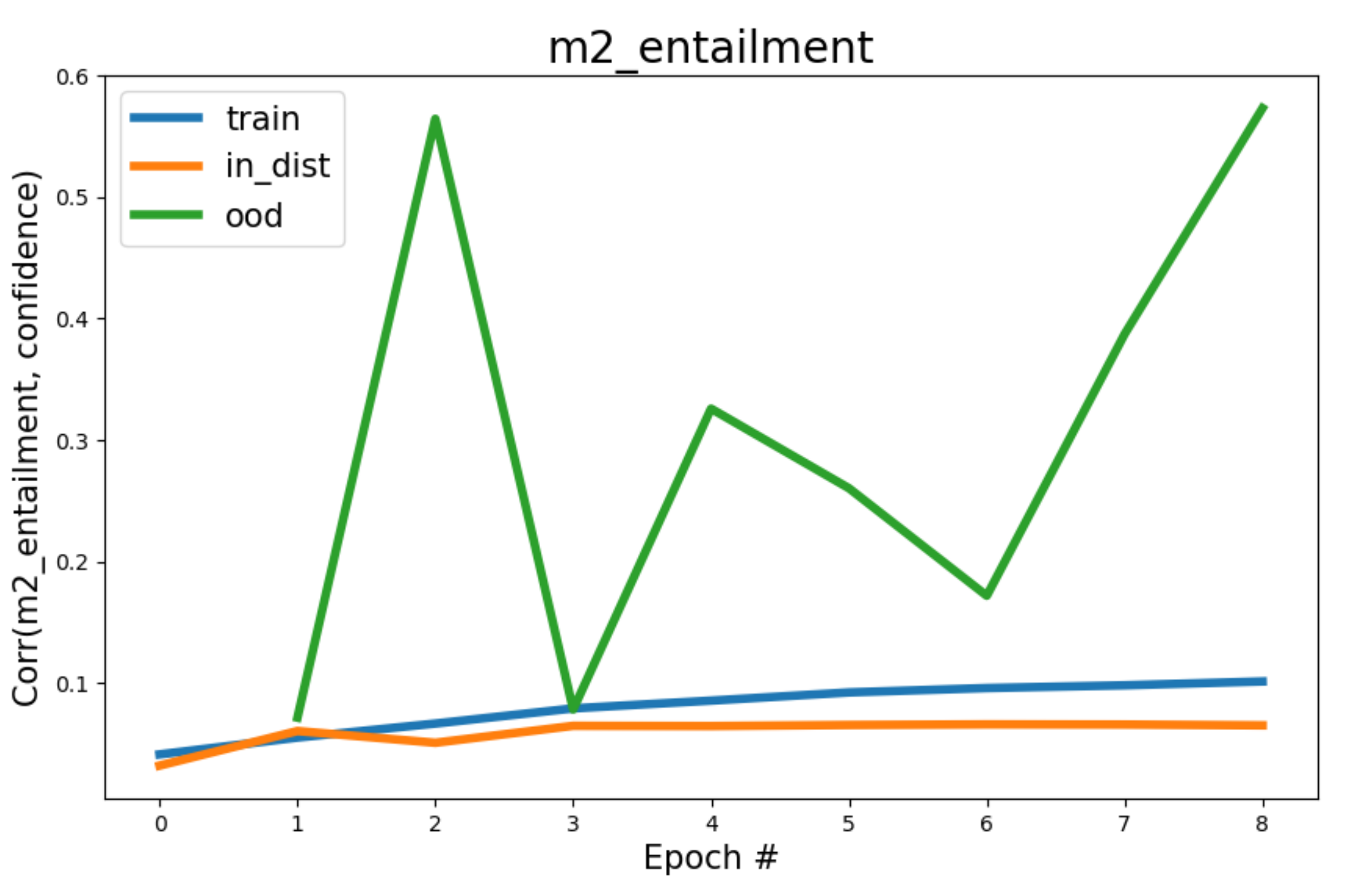}
    \caption{Correlation in \textbf{entailment} samples}
  \end{subfigure}
  \caption{Hypothesis \ref{sec:hypothesis-heuristic}: $\rho(m2, {\mu}_i^{E})$, $\rho(m2, {\mu}_{j(in-dist)}^{E})$, $\rho(m2, {\mu}_{j(OOD)}^{E})$ on \textit{all} samples (a) and \textit{entailment} samples(b)}
  \label{fig:result-ex2}
  \vspace{-10mm}
\end{wrapfigure}

For the second hypothesis, we find the following observations (refer to figure \ref{fig:result-ex2}):
\begin{itemize}[leftmargin=*]
    \item \textbf{Observation:} Correlation values ($\rho(m2, {\mu}_i^{E})$, $\rho(m2, {\mu}_{j(in-dist)}^{E})$, $\rho(m2, {\mu}_{j(OOD)}^{E})$) are relatively low in the plot for all samples (a). This applies to all cases (train, in-distribution, and OOD test). However, \textit{an obvious divergence is observed in the $\rho(m2, {\mu}_{j(OOD)}^{E})$ plot for the entailment samples (b)}.
    \\
    \textbf{Conjecture:} This observation confirms our hypothesis in \ref{sec:hypothesis-heuristic} that models are \textit{more prone to adopt syntactic heuristics when making inference on OOD samples}.
    \item \textbf{Observation:} All samples' trends adhere to the entailment trends in the train and in-distribution test case (orange and blue lines in figures (a) and (b)) but contradict the OOD case (green lines in figures (a) and (b)). In the OOD case, we can see that the negative and non-zero slopes are only found in all samples trend, but the correlation keeps increasing (with a huge difference in value) in the entailment samples trend. 
    \\
    \textbf{Conjecture:} This might indicate that knowledge used by the model to infer class labels is \textit{consistent for all classes in train and in-distribution but conflicts in the OOD case}. 
\end{itemize}


More supplementary plots of this analysis (more experiments on different datasets and the trend plots for \textbf{non-entailment} samples) can be found in appendix \ref{appendix:exp-2-non-entailment}.

\section{Limitations and Future Directions}
This data-centric approach to understanding the difference between OOD and in-distribution samples offers an intuition of failure mode in OOD. However, the results and approach presented in this paper have some limitations.

Although all NLI sets in our experiments came from different topic domains and had distinct characteristics, a more careful selection is needed to ensure maximum separation of in-distribution vs. OOD data distribution. Zhang et al. (2021) \cite{zhang2021understanding} highlights the problem with experiments setup where OOD regions can lie in support of the data distribution: high logit values assigned to certain OOD samples due to model estimation error. By ensuring this criterion is met, we can achieve a more accurate confidence score, which in turn will improve our results in section \ref{sec:result-ex1}.

More controlled experiment is needed for results in section \ref{sec:result-ex2} to ensure that the correlation between confidence and m2 is not spurious (i.e., there exists no unobserved confounding variable). One way to remedy this is by running the same measurements on random subsets and see if the correlation pattern still holds (i.e., randomized controlled experiments).

Future work may include remedies for the two mentioned limitations. Furthermore, investigating the root cause of contradicting syntactic heuristics of OOD and in-distribution samples that the model fails to predict correctly (section \ref{sec:result-ex1}) is also an exciting direction. It will also be exciting to see the effect of correcting for distribution mismatches \cite{schneider2020improving, fuchs2021distribution, wang2020tent} to the observations presented in this paper. To further corroborate the analysis and conjecture presented in this paper, the two possible directions can be: (i) adding more syntactic heuristics for analysis (e.g., coreference, negation), and (ii) setting up similar experiments on vision and other domains datasets. Finally, we hope that the insights provided by this data-centric view can motivate more informed development of OOD detection and generalization methods.

\bibliographystyle{plain}
\bibliography{main}

\begin{thebibliography}{10}

\bibitem{agrawal-etal-2016-analyzing}
Aishwarya Agrawal, Dhruv Batra, and Devi Parikh.
\newblock Analyzing the behavior of visual question answering models.
\newblock In {\em Proceedings of the 2016 Conference on Empirical Methods in
  Natural Language Processing}, pages 1955--1960, Austin, Texas, November 2016.
  Association for Computational Linguistics.

\bibitem{10.1007/978-3-030-01270-0_28}
Sara Beery, Grant Van~Horn, and Pietro Perona.
\newblock Recognition in terra incognita.
\newblock In Vittorio Ferrari, Martial Hebert, Cristian Sminchisescu, and Yair
  Weiss, editors, {\em Computer Vision -- ECCV 2018}, pages 472--489, Cham,
  2018. Springer International Publishing.

\bibitem{bendale2015towards}
Abhijit Bendale and Terrance Boult.
\newblock Towards open world recognition.
\newblock In {\em Proceedings of the IEEE conference on computer vision and
  pattern recognition}, pages 1893--1902, 2015.

\bibitem{bowman-etal-2015-large}
Samuel~R. Bowman, Gabor Angeli, Christopher Potts, and Christopher~D. Manning.
\newblock A large annotated corpus for learning natural language inference.
\newblock In {\em Proceedings of the 2015 Conference on Empirical Methods in
  Natural Language Processing}, pages 632--642, Lisbon, Portugal, September
  2015. Association for Computational Linguistics.

\bibitem{chen2021robustifying}
Jiefeng Chen, Yixuan Li, Xi~Wu, Yingyu Liang, and Somesh Jha.
\newblock Atom: Robustifying out-of-distribution detection using outlier
  mining.
\newblock {\em In Proceedings of European Conference on Machine Learning and
  Principles and Practice of Knowledge Discovery in Databases (ECML PKDD)},
  2021.

\bibitem{Choi2018WAICBW}
Hyun-Jae Choi, Eric Jang, and Alexander~Amir Alemi.
\newblock Waic, but why? generative ensembles for robust anomaly detection.
\newblock {\em arXiv: Machine Learning}, 2018.

\bibitem{condoravdi-etal-2003-entailment}
Cleo Condoravdi, Dick Crouch, Valeria de~Paiva, Reinhard Stolle, and Daniel~G.
  Bobrow.
\newblock Entailment, intensionality and text understanding.
\newblock In {\em Proceedings of the {HLT}-{NAACL} 2003 Workshop on Text
  Meaning}, pages 38--45, 2003.

\bibitem{10.1007/11736790_9}
Ido Dagan, Oren Glickman, and Bernardo Magnini.
\newblock The pascal recognising textual entailment challenge.
\newblock In Joaquin Qui{\~{n}}onero-Candela, Ido Dagan, Bernardo Magnini, and
  Florence d'Alch{\'e} Buc, editors, {\em Machine Learning Challenges.
  Evaluating Predictive Uncertainty, Visual Object Classification, and
  Recognising Tectual Entailment}, pages 177--190, Berlin, Heidelberg, 2006.
  Springer Berlin Heidelberg.

\bibitem{fetaya2019understanding}
Ethan Fetaya, J{\"o}rn-Henrik Jacobsen, Will Grathwohl, and Richard Zemel.
\newblock Understanding the limitations of conditional generative models.
\newblock {\em arXiv preprint arXiv:1906.01171}, 2019.

\bibitem{fuchs2021distribution}
Alexander Fuchs, Christian Knoll, and Franz Pernkopf.
\newblock Distribution mismatch correction for improved robustness in deep
  neural networks.
\newblock {\em arXiv preprint arXiv:2110.01955}, 2021.

\bibitem{hendrycks17baseline}
Dan Hendrycks and Kevin Gimpel.
\newblock A baseline for detecting misclassified and out-of-distribution
  examples in neural networks.
\newblock {\em Proceedings of International Conference on Learning
  Representations}, 2017.

\bibitem{hendrycks-etal-2020-pretrained}
Dan Hendrycks, Xiaoyuan Liu, Eric Wallace, Adam Dziedzic, Rishabh Krishnan, and
  Dawn Song.
\newblock Pretrained transformers improve out-of-distribution robustness.
\newblock In {\em Proceedings of the 58th Annual Meeting of the Association for
  Computational Linguistics}, pages 2744--2751, Online, July 2020. Association
  for Computational Linguistics.

\bibitem{jiang2021x}
Jingjing Jiang, Ziyi Liu, Yifan Liu, Zhixiong Nan, and Nanning Zheng.
\newblock X-ggm: Graph generative modeling for out-of-distribution
  generalization in visual question answering.
\newblock In {\em Proceedings of the 29th ACM International Conference on
  Multimedia}, pages 199--208, 2021.

\bibitem{just2019deep}
John Just and Sambuddha Ghosal.
\newblock Deep generative models strike back! improving understanding and
  evaluation in light of unmet expectations for ood data.
\newblock {\em arXiv preprint arXiv:1911.04699}, 2019.

\bibitem{kirichenko2020normalizing}
Polina Kirichenko, Pavel Izmailov, and Andrew~Gordon Wilson.
\newblock Why normalizing flows fail to detect out-of-distribution data, 2020.

\bibitem{10.5555/3295222.3295387}
Balaji Lakshminarayanan, Alexander Pritzel, and Charles Blundell.
\newblock Simple and scalable predictive uncertainty estimation using deep
  ensembles.
\newblock In {\em Proceedings of the 31st International Conference on Neural
  Information Processing Systems}, NIPS'17, page 6405–6416, Red Hook, NY,
  USA, 2017. Curran Associates Inc.

\bibitem{lan2020perfect}
Charline~Le Lan and Laurent Dinh.
\newblock Perfect density models cannot guarantee anomaly detection.
\newblock {\em arXiv preprint arXiv:2012.03808}, 2020.

\bibitem{lee2018simple}
Kimin Lee, Kibok Lee, Honglak Lee, and Jinwoo Shin.
\newblock A simple unified framework for detecting out-of-distribution samples
  and adversarial attacks.
\newblock {\em Advances in neural information processing systems}, 31, 2018.

\bibitem{levesque2012winograd}
Hector Levesque, Ernest Davis, and Leora Morgenstern.
\newblock The winograd schema challenge.
\newblock In {\em Thirteenth International Conference on the Principles of
  Knowledge Representation and Reasoning}, 2012.

\bibitem{liu2020energy}
Weitang Liu, Xiaoyun Wang, John~D Owens, and Yixuan Li.
\newblock Energy-based out-of-distribution detection.
\newblock {\em arXiv preprint arXiv:2010.03759}, 2020.

\bibitem{liu2019roberta}
Yinhan Liu, Myle Ott, Naman Goyal, Jingfei Du, Mandar Joshi, Danqi Chen, Omer
  Levy, Mike Lewis, Luke Zettlemoyer, and Veselin Stoyanov.
\newblock Roberta: A robustly optimized bert pretraining approach.
\newblock {\em arXiv preprint arXiv:1907.11692}, 2019.

\bibitem{mccoy-etal-2019-right}
Tom McCoy, Ellie Pavlick, and Tal Linzen.
\newblock Right for the wrong reasons: Diagnosing syntactic heuristics in
  natural language inference.
\newblock In {\em Proceedings of the 57th Annual Meeting of the Association for
  Computational Linguistics}, pages 3428--3448, Florence, Italy, July 2019.
  Association for Computational Linguistics.

\bibitem{mohseni2020self}
Sina Mohseni, Mandar Pitale, JBS Yadawa, and Zhangyang Wang.
\newblock Self-supervised learning for generalizable out-of-distribution
  detection.
\newblock In {\em Proceedings of the AAAI Conference on Artificial
  Intelligence}, volume~34, pages 5216--5223, 2020.

\bibitem{naik-etal-2018-stress}
Aakanksha Naik, Abhilasha Ravichander, Norman Sadeh, Carolyn Rose, and Graham
  Neubig.
\newblock Stress test evaluation for natural language inference.
\newblock In {\em Proceedings of the 27th International Conference on
  Computational Linguistics}, pages 2340--2353, Santa Fe, New Mexico, USA,
  August 2018. Association for Computational Linguistics.

\bibitem{pytorch}
Adam Paszke, Sam Gross, Francisco Massa, Adam Lerer, James Bradbury, Gregory
  Chanan, Trevor Killeen, Zeming Lin, Natalia Gimelshein, Luca Antiga, Alban
  Desmaison, Andreas Kopf, Edward Yang, Zachary DeVito, Martin Raison, Alykhan
  Tejani, Sasank Chilamkurthy, Benoit Steiner, Lu~Fang, Junjie Bai, and Soumith
  Chintala.
\newblock Pytorch: An imperative style, high-performance deep learning library.
\newblock In H.~Wallach, H.~Larochelle, A.~Beygelzimer, F.~d\textquotesingle
  Alch\'{e}-Buc, E.~Fox, and R.~Garnett, editors, {\em Advances in Neural
  Information Processing Systems 32}, pages 8024--8035. Curran Associates,
  Inc., 2019.

\bibitem{sanchez-etal-2018-behavior}
Ivan Sanchez, Jeff Mitchell, and Sebastian Riedel.
\newblock Behavior analysis of {NLI} models: Uncovering the influence of three
  factors on robustness.
\newblock In {\em Proceedings of the 2018 Conference of the North {A}merican
  Chapter of the Association for Computational Linguistics: Human Language
  Technologies, Volume 1 (Long Papers)}, pages 1975--1985, New Orleans,
  Louisiana, June 2018. Association for Computational Linguistics.

\bibitem{schneider2020improving}
Steffen Schneider, Evgenia Rusak, Luisa Eck, Oliver Bringmann, Wieland Brendel,
  and Matthias Bethge.
\newblock Improving robustness against common corruptions by covariate shift
  adaptation.
\newblock {\em Advances in Neural Information Processing Systems}, 33, 2020.

\bibitem{swayamdipta-etal-2020-dataset}
Swabha Swayamdipta, Roy Schwartz, Nicholas Lourie, Yizhong Wang, Hannaneh
  Hajishirzi, Noah~A. Smith, and Yejin Choi.
\newblock Dataset cartography: Mapping and diagnosing datasets with training
  dynamics.
\newblock In {\em Proceedings of the 2020 Conference on Empirical Methods in
  Natural Language Processing (EMNLP)}, pages 9275--9293, Online, November
  2020. Association for Computational Linguistics.

\bibitem{tayal2020model}
Kshitij Tayal, Rahul Ghosh, and Vipin Kumar.
\newblock Model-agnostic methods for text classification with inherent noise.
\newblock In {\em Proceedings of the 28th International Conference on
  Computational Linguistics: Industry Track}, pages 202--213, 2020.

\bibitem{tayal2020regularized}
Kshitij Tayal, Nikhil Rao, Saurabh Agarwal, Xiaowei Jia, Karthik Subbian, and
  Vipin Kumar.
\newblock Regularized graph convolutional networks for short text
  classification.
\newblock In {\em Proceedings of the 28th International Conference on
  Computational Linguistics: Industry Track}, pages 236--242, 2020.

\bibitem{wang2019glue}
Alex Wang, Amanpreet Singh, Julian Michael, Felix Hill, Omer Levy, and
  Samuel~R. Bowman.
\newblock {GLUE}: A multi-task benchmark and analysis platform for natural
  language understanding.
\newblock 2019.
\newblock In the Proceedings of ICLR.

\bibitem{wang2020tent}
Dequan Wang, Evan Shelhamer, Shaoteng Liu, Bruno Olshausen, and Trevor Darrell.
\newblock Tent: Fully test-time adaptation by entropy minimization.
\newblock {\em arXiv preprint arXiv:2006.10726}, 2020.

\bibitem{wang2017visual}
Jianyu {Wang}, Zhishuai {Zhang}, Cihang {Xie}, Yuyin {Zhou}, Vittal
  {Premachandran}, Jun {Zhu}, Lingxi {Xie}, and Alan {Yuille}.
\newblock Visual concepts and compositional voting.
\newblock {\em arXiv preprint arXiv:1711.04451}, 2017.

\bibitem{wang2020analysis}
Ziyu Wang, Bin Dai, David Wipf, and Jun Zhu.
\newblock Further analysis of outlier detection with deep generative models,
  2020.

\bibitem{williams-etal-2018-broad}
Adina Williams, Nikita Nangia, and Samuel Bowman.
\newblock A broad-coverage challenge corpus for sentence understanding through
  inference.
\newblock In {\em Proceedings of the 2018 Conference of the North {A}merican
  Chapter of the Association for Computational Linguistics: Human Language
  Technologies, Volume 1 (Long Papers)}, pages 1112--1122, New Orleans,
  Louisiana, June 2018. Association for Computational Linguistics.

\bibitem{10.1007/978-3-030-58580-8_7}
Hongjie Zhang, Ang Li, Jie Guo, and Yanwen Guo.
\newblock Hybrid models for open set recognition.
\newblock In Andrea Vedaldi, Horst Bischof, Thomas Brox, and Jan-Michael Frahm,
  editors, {\em Computer Vision -- ECCV 2020}, pages 102--117, Cham, 2020.
  Springer International Publishing.

\bibitem{zhang2021understanding}
Lily Zhang, Mark Goldstein, and Rajesh Ranganath.
\newblock Understanding failures in out-of-distribution detection with deep
  generative models.
\newblock In {\em International Conference on Machine Learning}, pages
  12427--12436. PMLR, 2021.

\end{thebibliography}

\newpage
\appendix

\section{Appendix}

\subsection{Supplementary plot:  OOD vs in-distribution on training dynamics information (Training and in-dis: RTE; OOD: WNLI)}
\label{appendix:exp-1-rte}
\begin{figure}[htbp]
  \begin{subfigure}[t]{.45\textwidth}
    \centering
    \includegraphics[width=\linewidth]{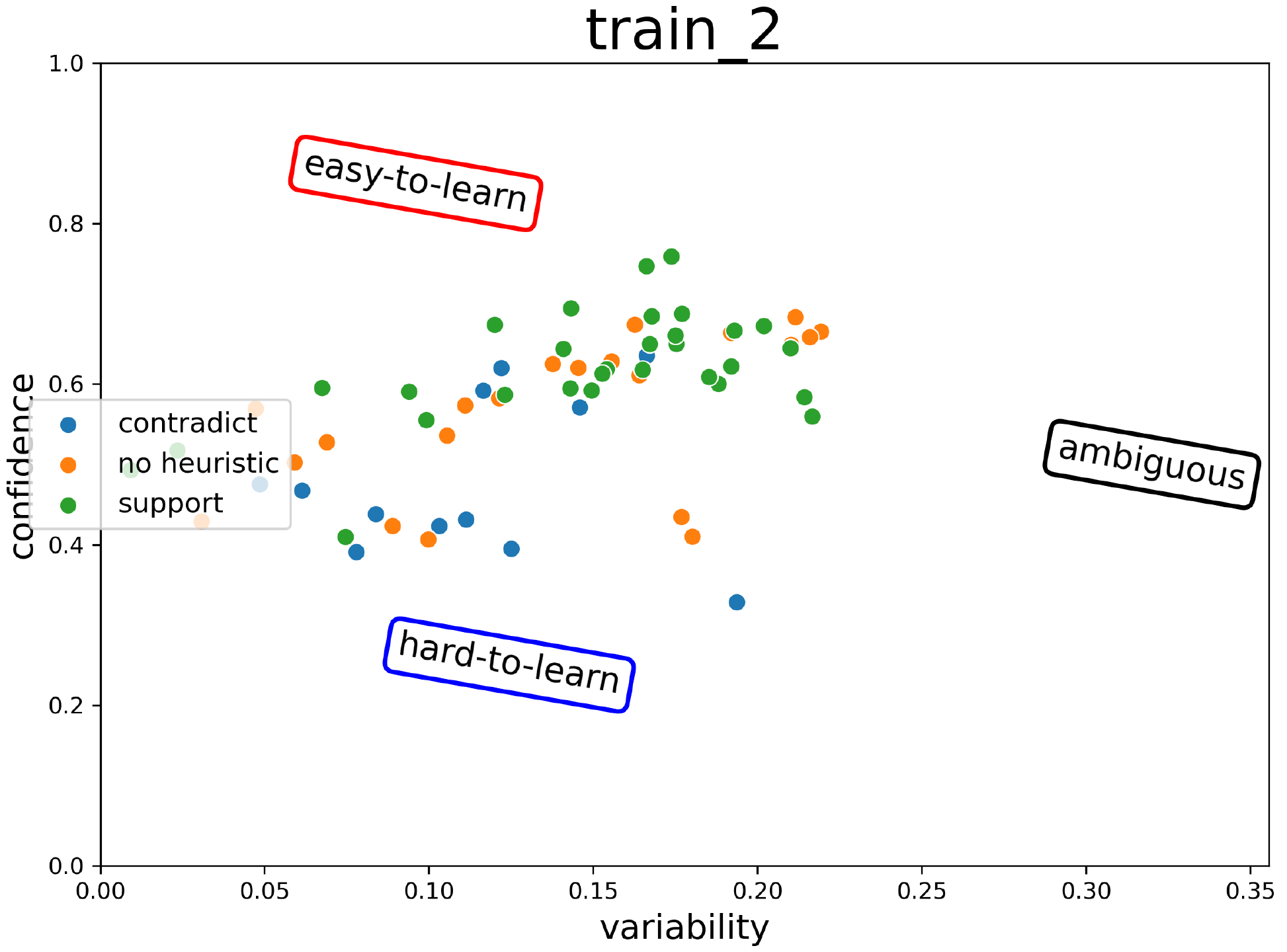}
    \caption{\textbf{Training} cartography map at \textbf{epoch 2}}
  \end{subfigure}
  \hfill
  \begin{subfigure}[t]{.45\textwidth}
    \centering
    \includegraphics[width=\linewidth]{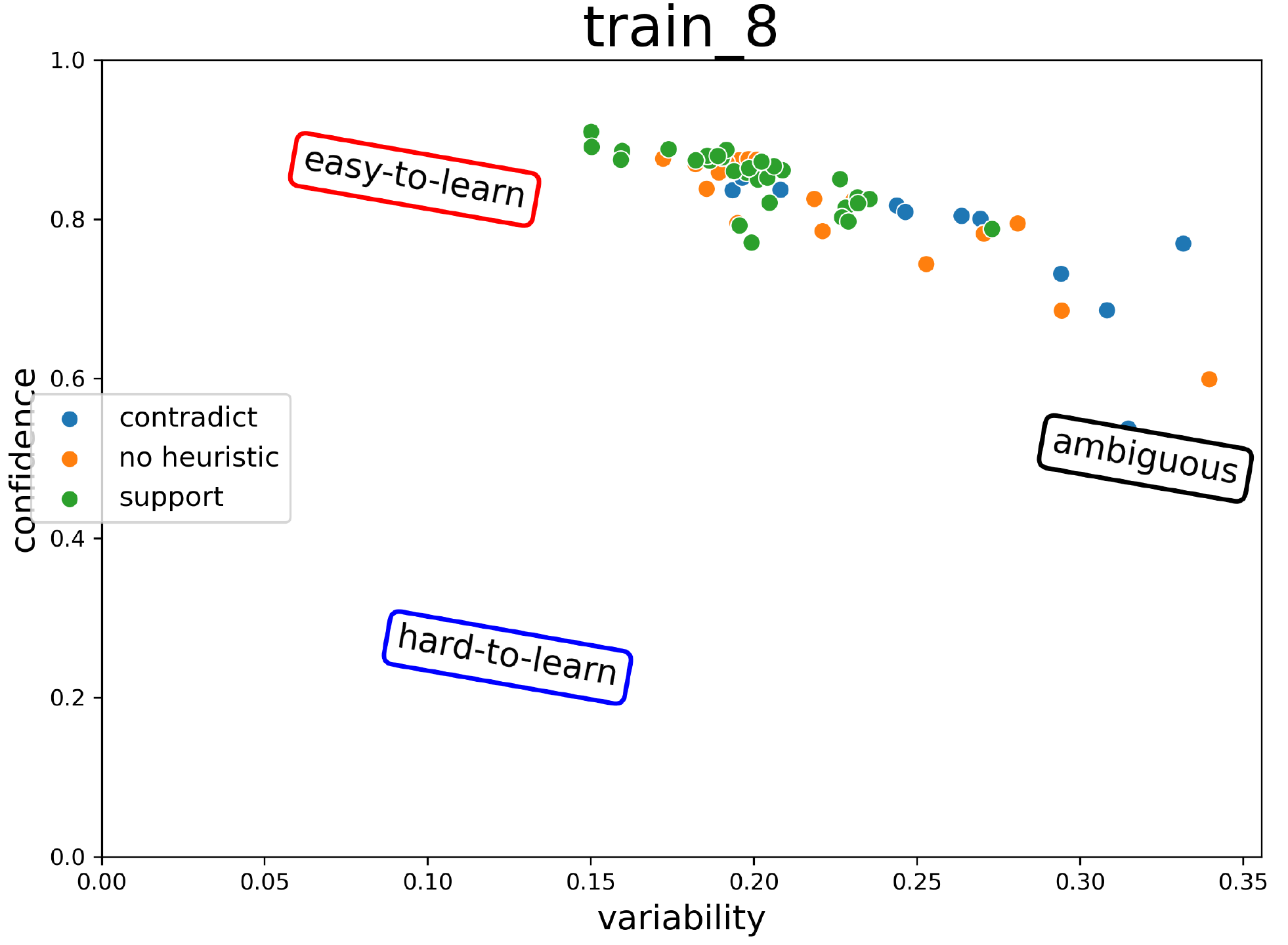}
    \caption{\textbf{Training} cartography map at \textbf{epoch 8}}
  \end{subfigure}
  \medskip
  \begin{subfigure}[t]{.45\textwidth}
    \centering
    \includegraphics[width=\linewidth]{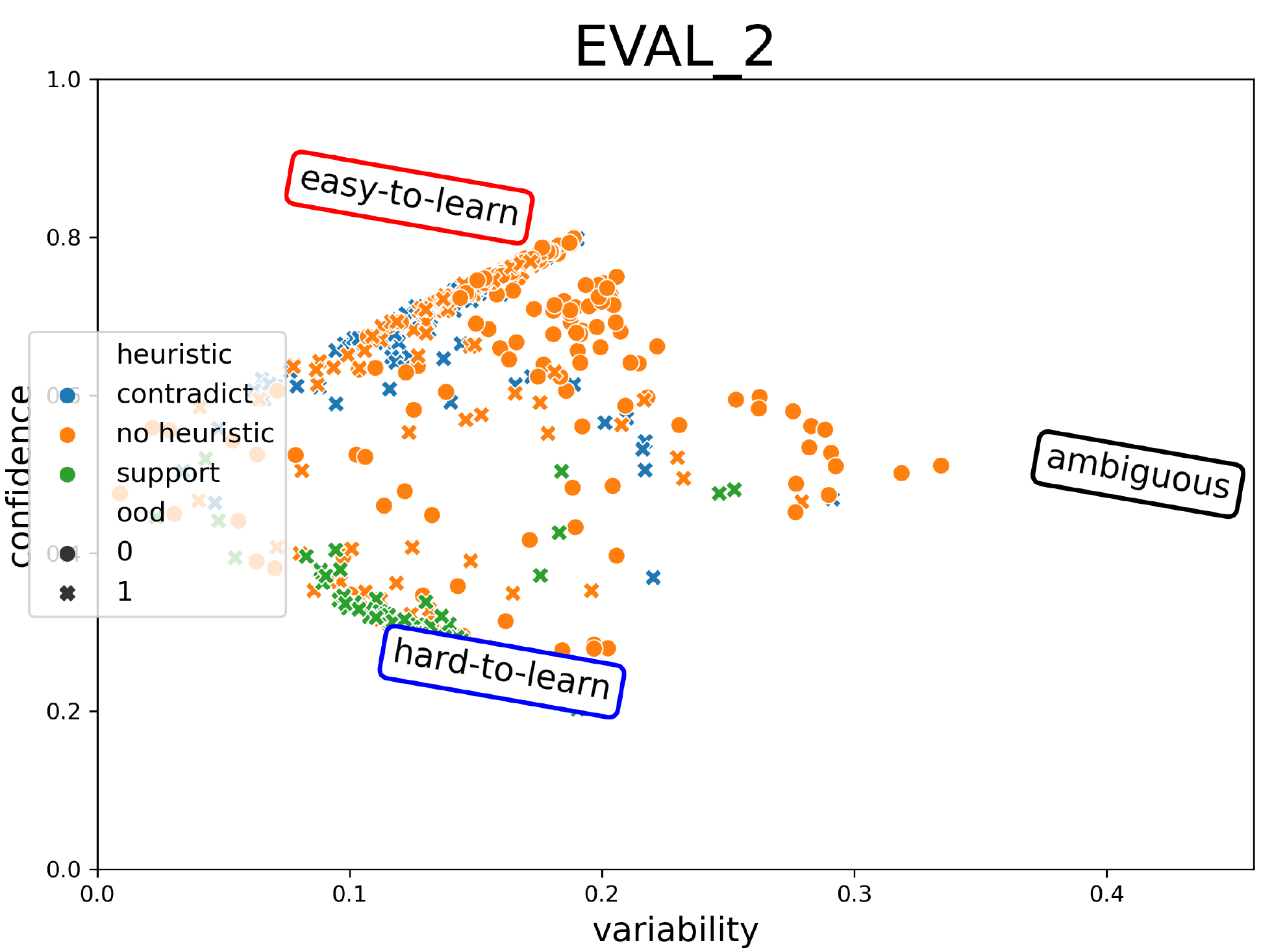}
    \caption{\textbf{Evaluation} cartography map at \textbf{epoch 2}}
  \end{subfigure}
  \hfill
  \begin{subfigure}[t]{.45\textwidth}
    \centering
    \includegraphics[width=\linewidth]{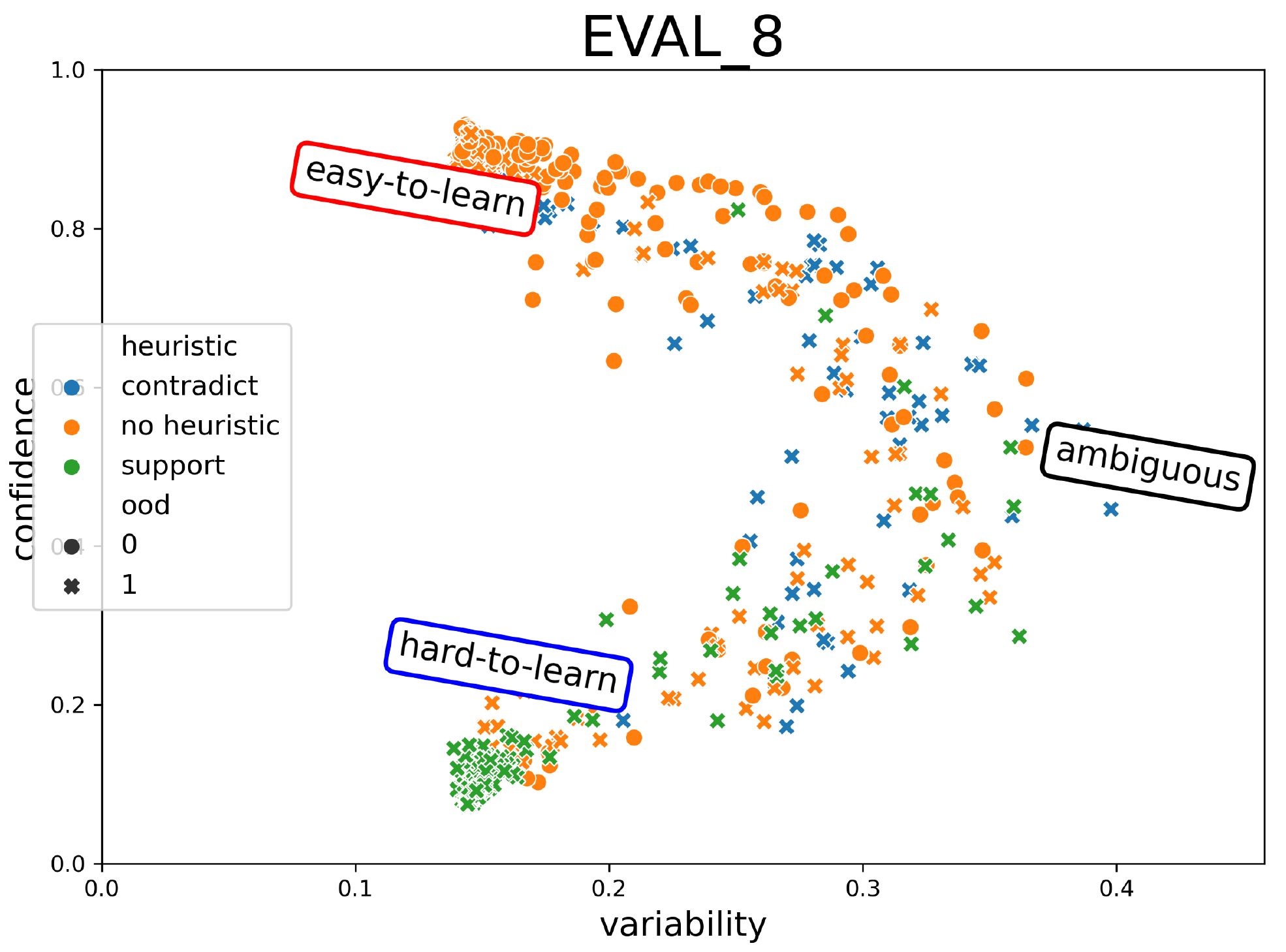}
    \caption{\textbf{Evaluation} cartography map at \textbf{epoch 8}}
  \end{subfigure}
 \caption{Training cartography maps (training set: RTE). The number of heuristics related samples in RTE is small.}
\end{figure}

\subsection{Supplementary plot:  OOD vs in-distribution on syntactic characteristics (entailment)}
\label{appendix:exp-2-entailment}
\begin{figure}[H]
  \begin{subfigure}[t]{.45\textwidth}
    \centering
    \includegraphics[width=\linewidth]{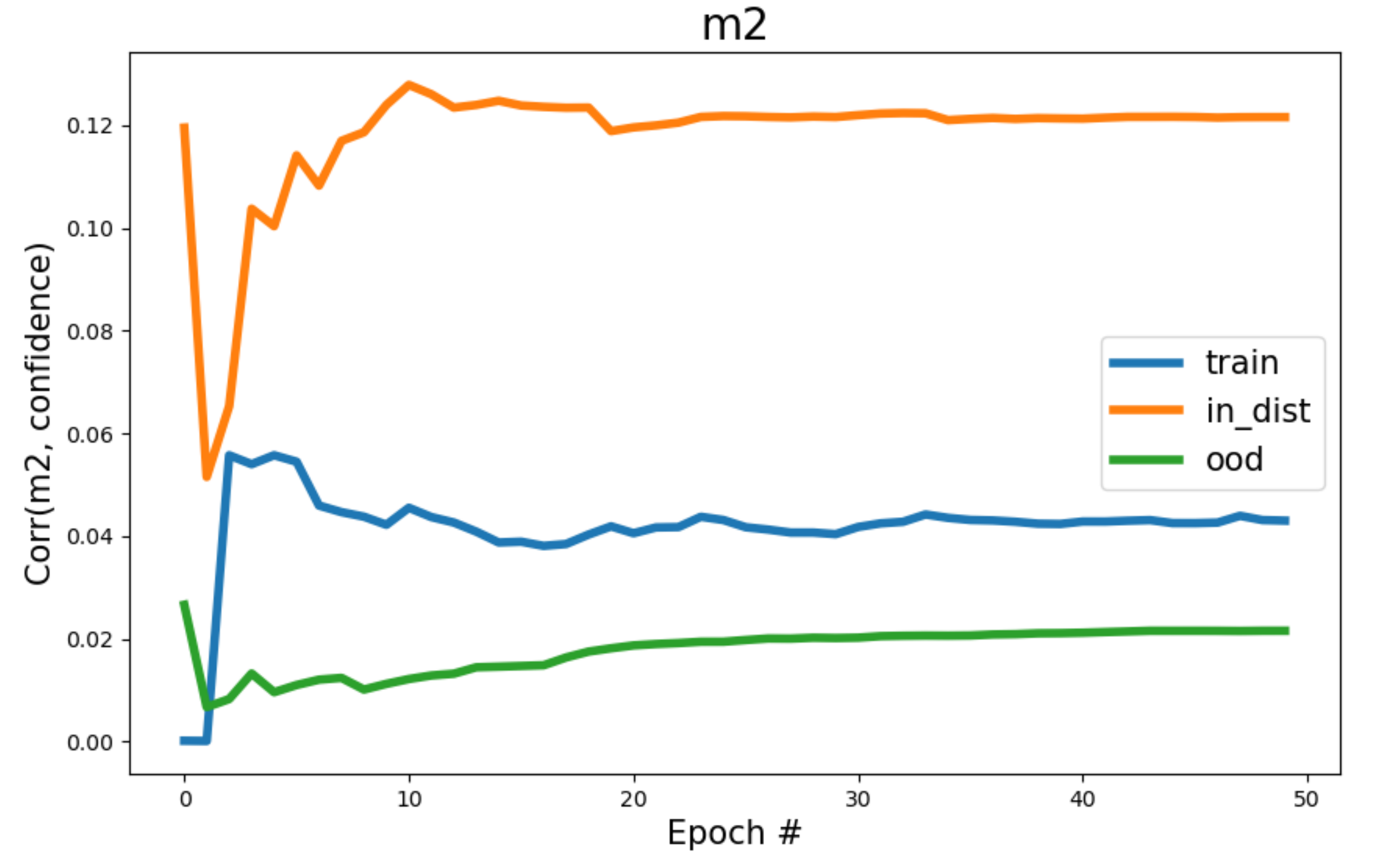}
    \caption{Correlation between \textbf{m2} and \textbf{all} samples $\hat{\mu}_i$}
  \end{subfigure}
  \hfill
  \begin{subfigure}[t]{.45\textwidth}
    \centering
    \includegraphics[width=\linewidth]{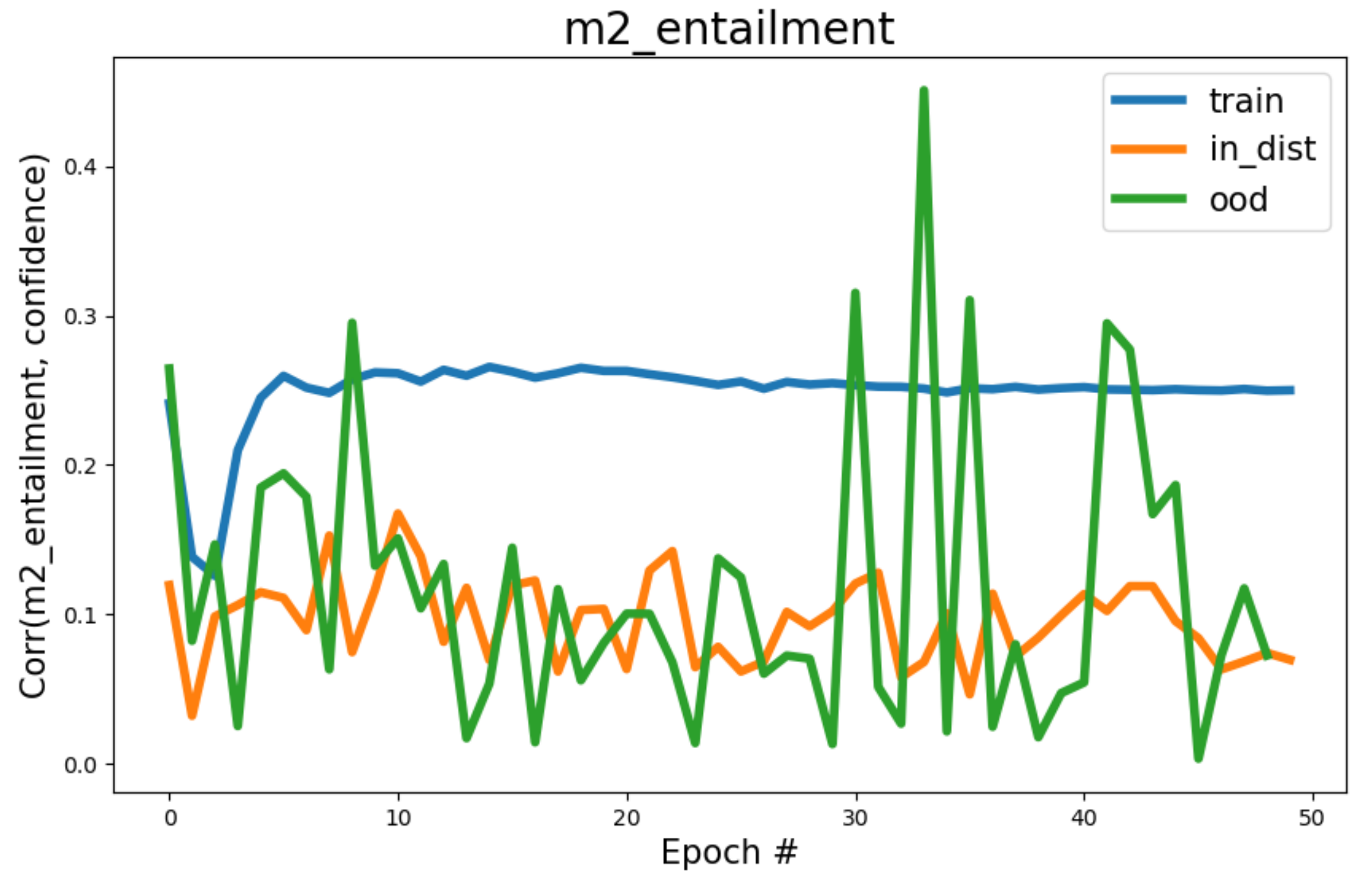}
    \caption{Correlation between \textbf{m2} and \textbf{entailment} samples $\hat{\mu}_i$}
  \end{subfigure}
  \caption{Results for hypothesis \ref{sec:hypothesis-heuristic}. Training and in-distribution test samples are RTE, and OOD samples are WNLI.}
\end{figure}

\begin{figure}[H]
  \begin{subfigure}[t]{.45\textwidth}
    \centering
    \includegraphics[width=\linewidth]{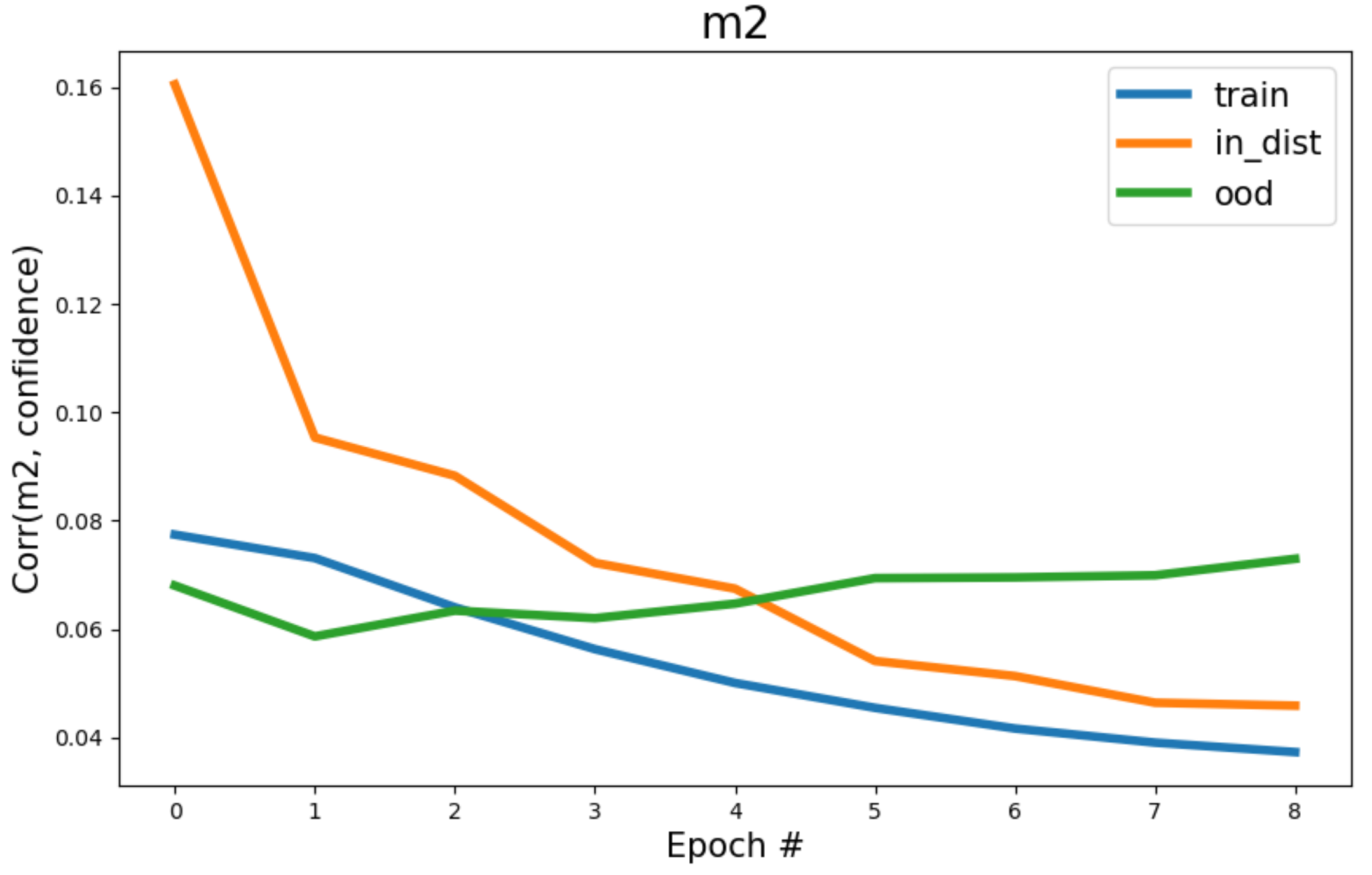}
    \caption{Correlation between \textbf{m2} and \textbf{all} samples $\hat{\mu}_i$}
  \end{subfigure}
  \hfill
  \begin{subfigure}[t]{.45\textwidth}
    \centering
    \includegraphics[width=\linewidth]{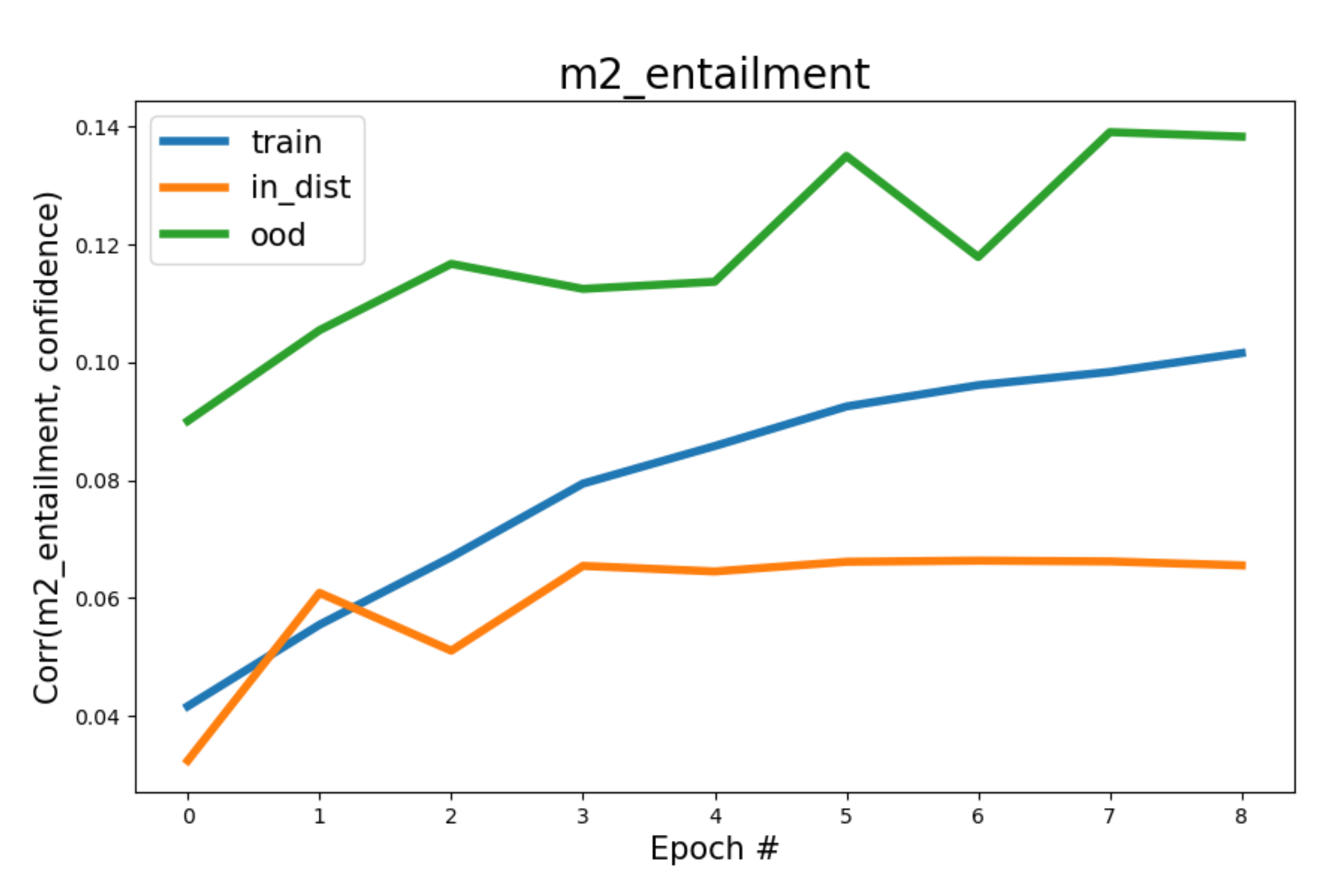}
    \caption{Correlation between \textbf{m2} and \textbf{entailment} samples $\hat{\mu}_i$}
  \end{subfigure}
  \caption{Results for hypothesis \ref{sec:hypothesis-heuristic}. Training and in-distribution test samples are MNLI, and OOD samples are RTE.}
\end{figure}
Results presented are at the end of epoch 8 for MNLI training and the end of epoch 50 for RTE training. This is based on the epoch in which the training error has converged (around 0.02).
\subsection{Supplementary plot:  OOD vs in-distribution on training dynamics information (Training and in-dis: MNLI; OOD: RTE)}
\label{appendix:exp-1-mnli}
\begin{figure}[htbp]
  \begin{subfigure}[t]{.45\textwidth}
    \centering
    \includegraphics[width=\linewidth]{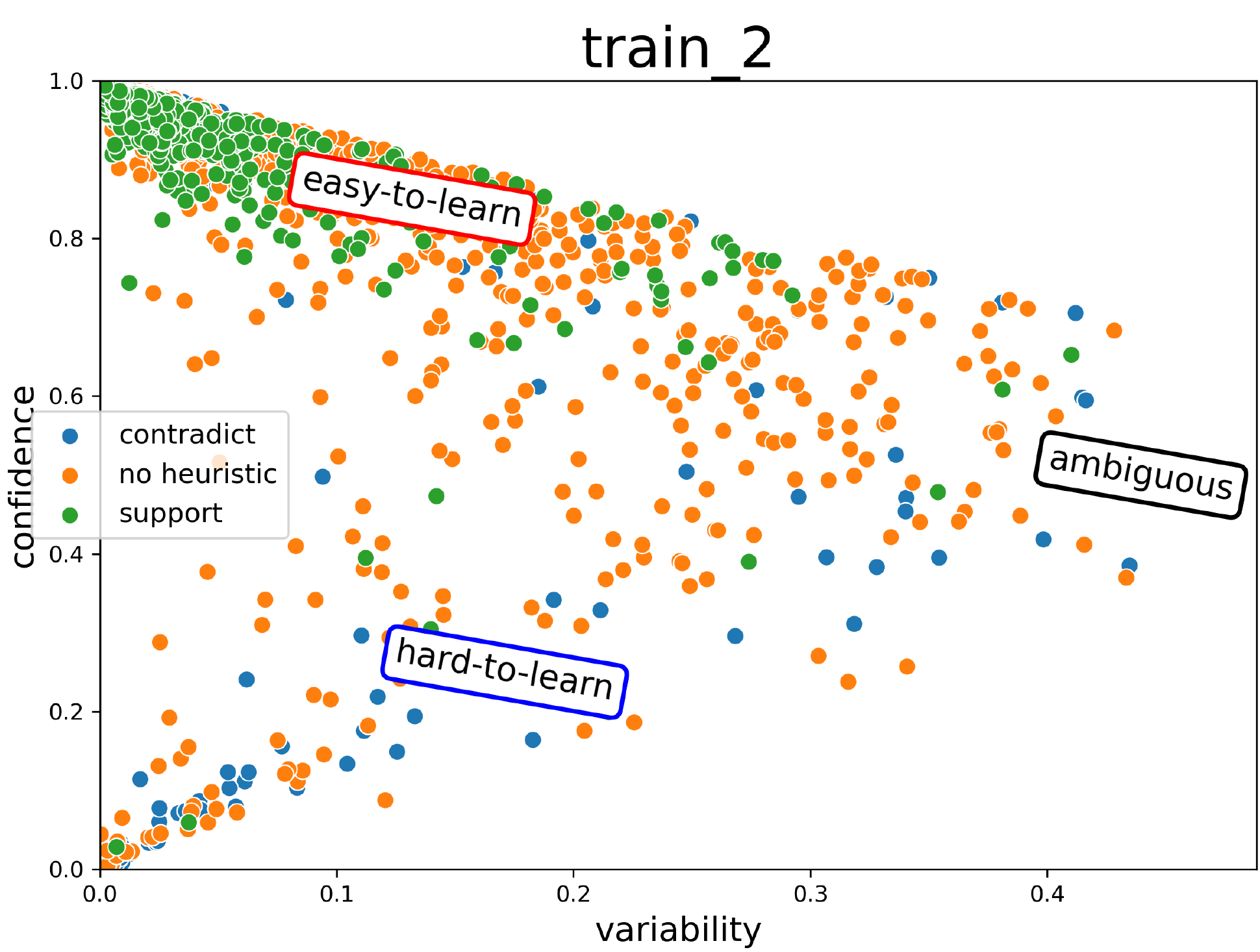}
    \caption{\textbf{Training} cartography map at \textbf{epoch 2}}
  \end{subfigure}
  \hfill
  \begin{subfigure}[t]{.45\textwidth}
    \centering
    \includegraphics[width=\linewidth]{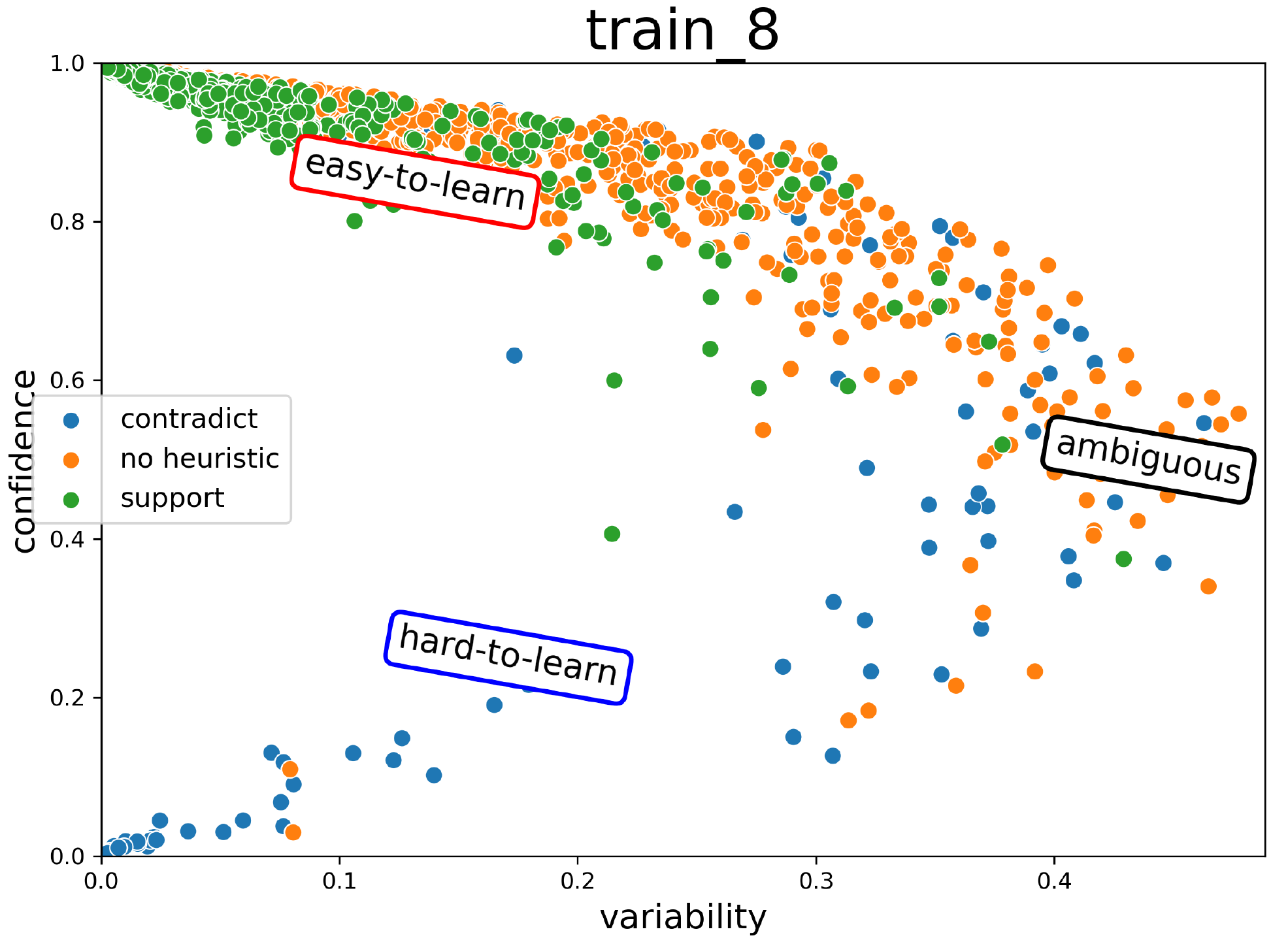}
    \caption{\textbf{Training} cartography map at \textbf{epoch 8}}
  \end{subfigure}
  \medskip
  \begin{subfigure}[t]{.45\textwidth}
    \centering
    \includegraphics[width=\linewidth]{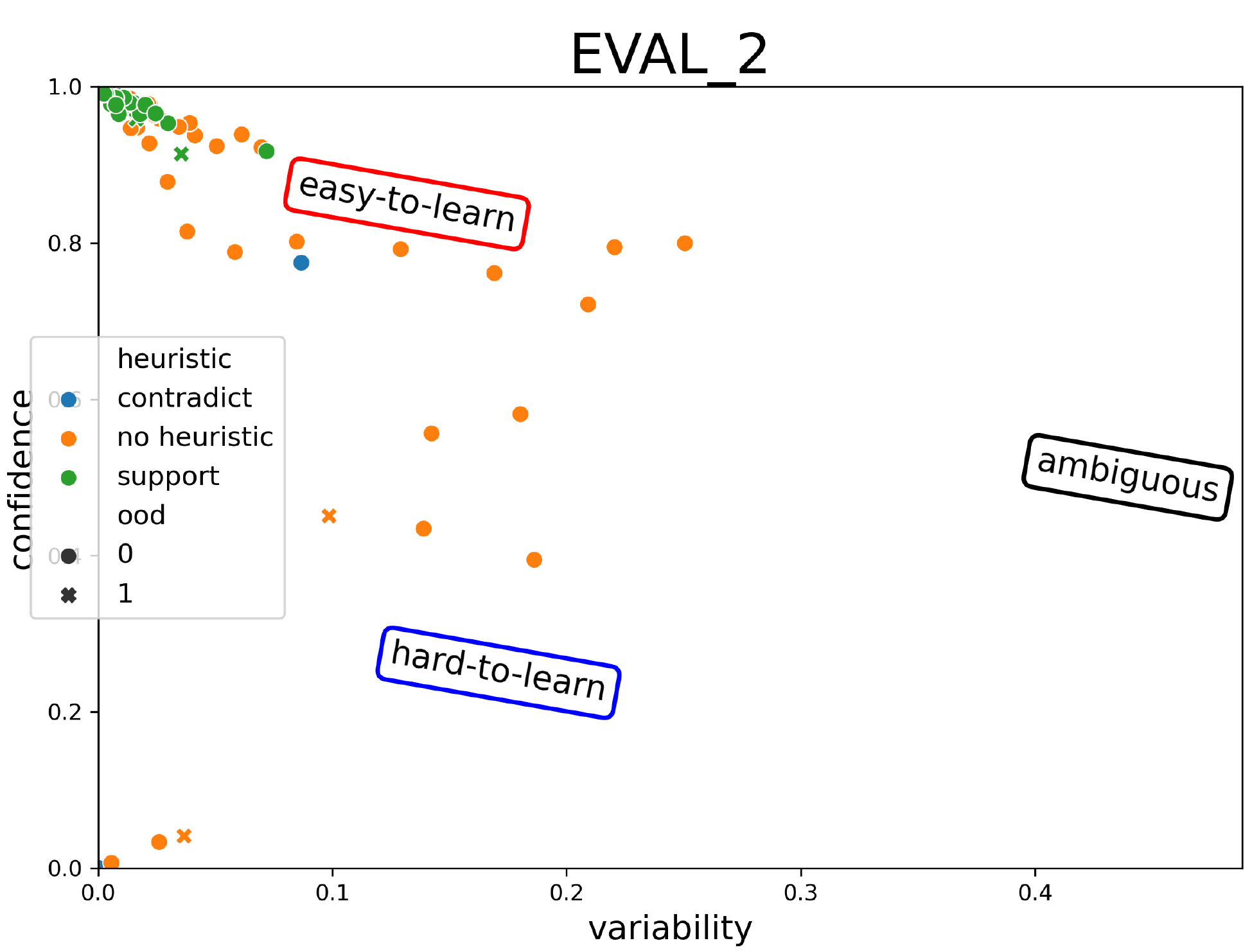}
    \caption{\textbf{Evaluation} cartography map at \textbf{epoch 2}}
  \end{subfigure}
  \hfill
  \begin{subfigure}[t]{.45\textwidth}
    \centering
    \includegraphics[width=\linewidth]{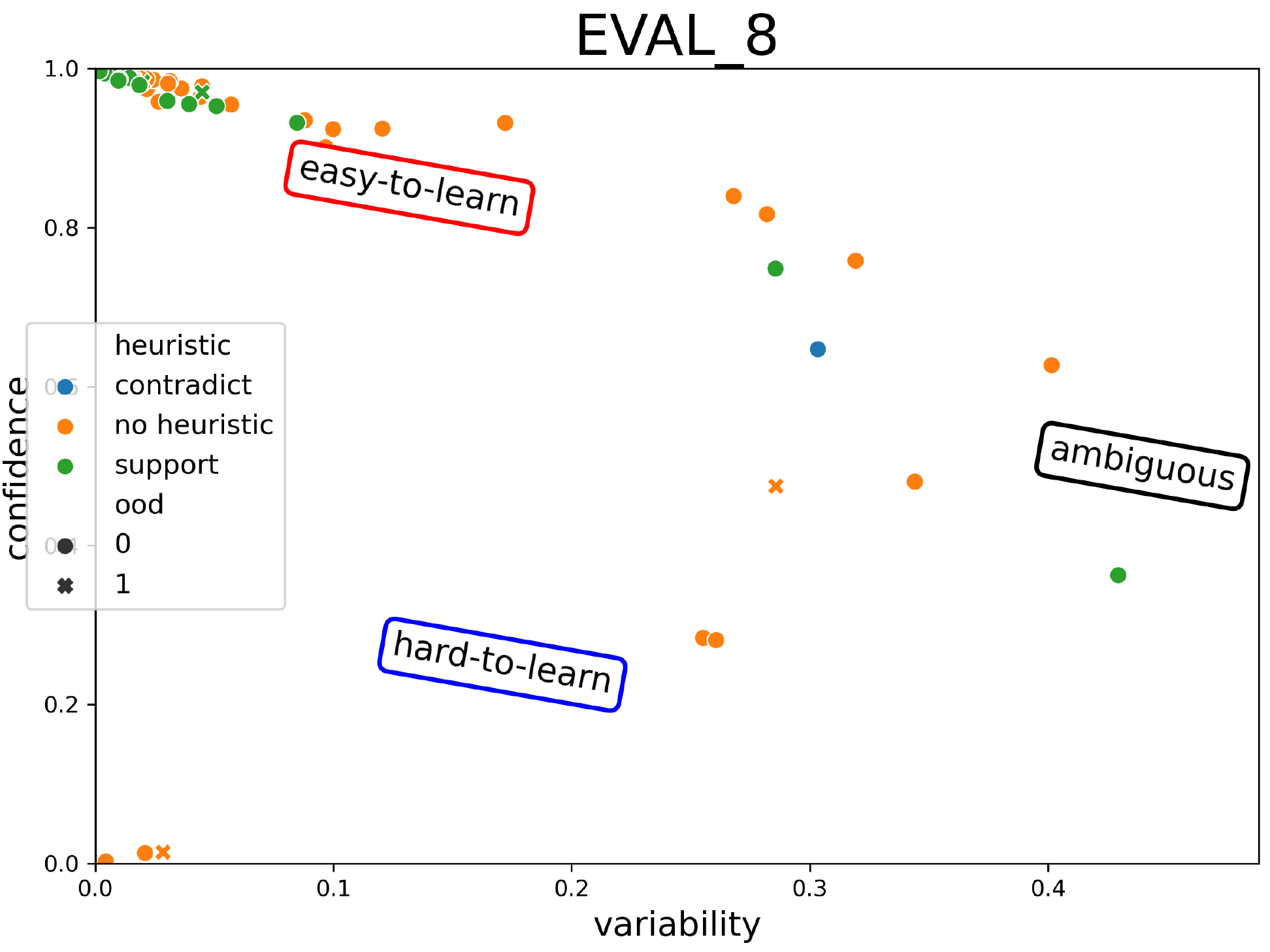}
    \caption{\textbf{Evaluation} cartography map at \textbf{epoch 8}}
  \end{subfigure}
 \caption{Training and evaluation cartography maps (train: MNLI, evaluation: RTE). The number of heuristics related samples in RTE is small.}
\end{figure}
\subsection{Supplementary plot:  OOD vs in-distribution on syntactic characteristics (non-entailment)}
\label{appendix:exp-2-non-entailment}
This section shows plots for correlation between confidence scores ($\hat{\mu}_i$) of \textbf{non entailment} samples and m2
\begin{figure}[H]
  \begin{subfigure}[t]{.32\textwidth}
    \centering
    \includegraphics[width=\linewidth]{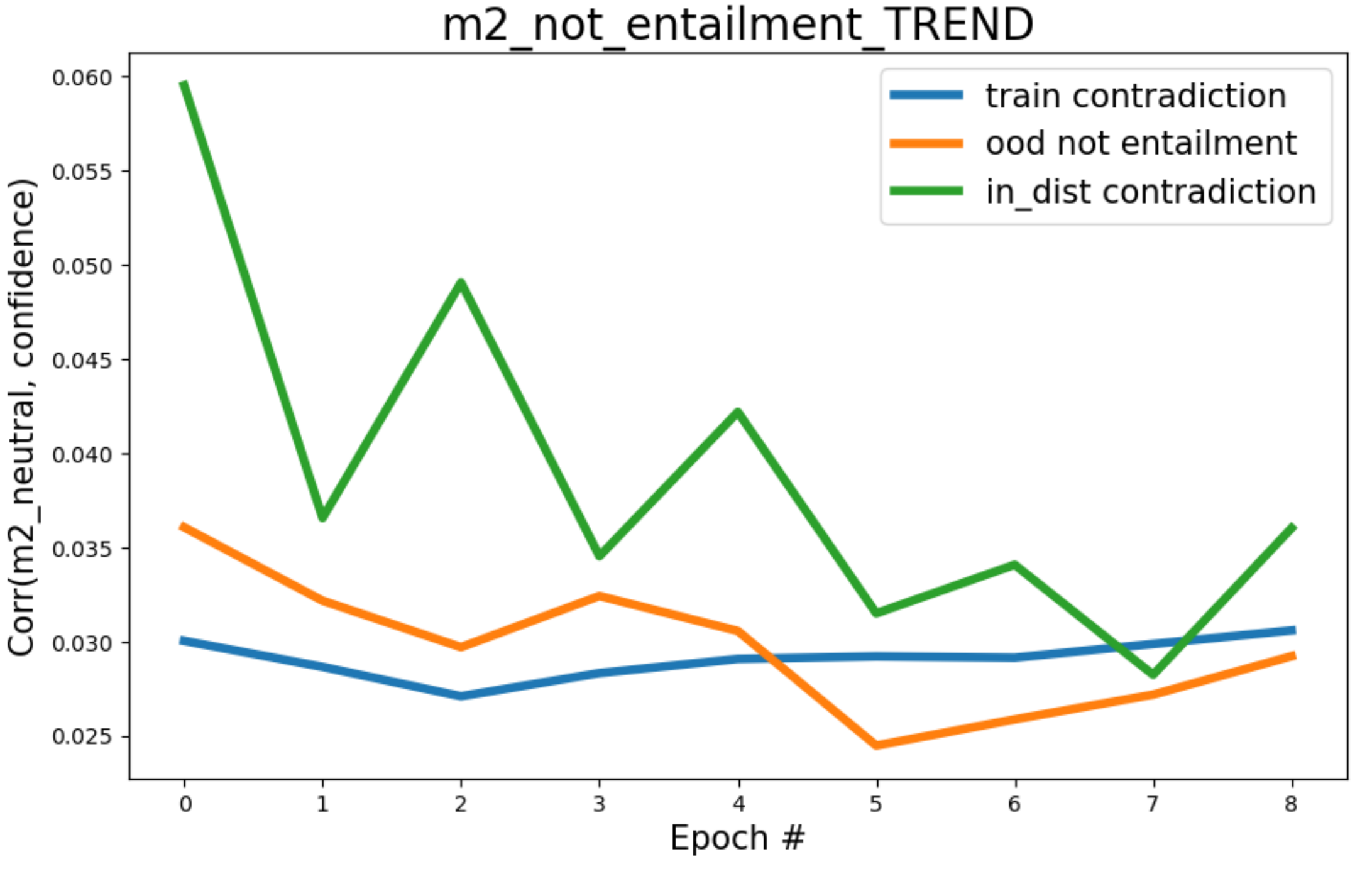}
    \caption{Train \& in-distribution: MNLI, OOD: WNLI}
  \end{subfigure}
   \hfill
    \begin{subfigure}[t]{.32\textwidth}
    \centering
    \includegraphics[width=\linewidth]{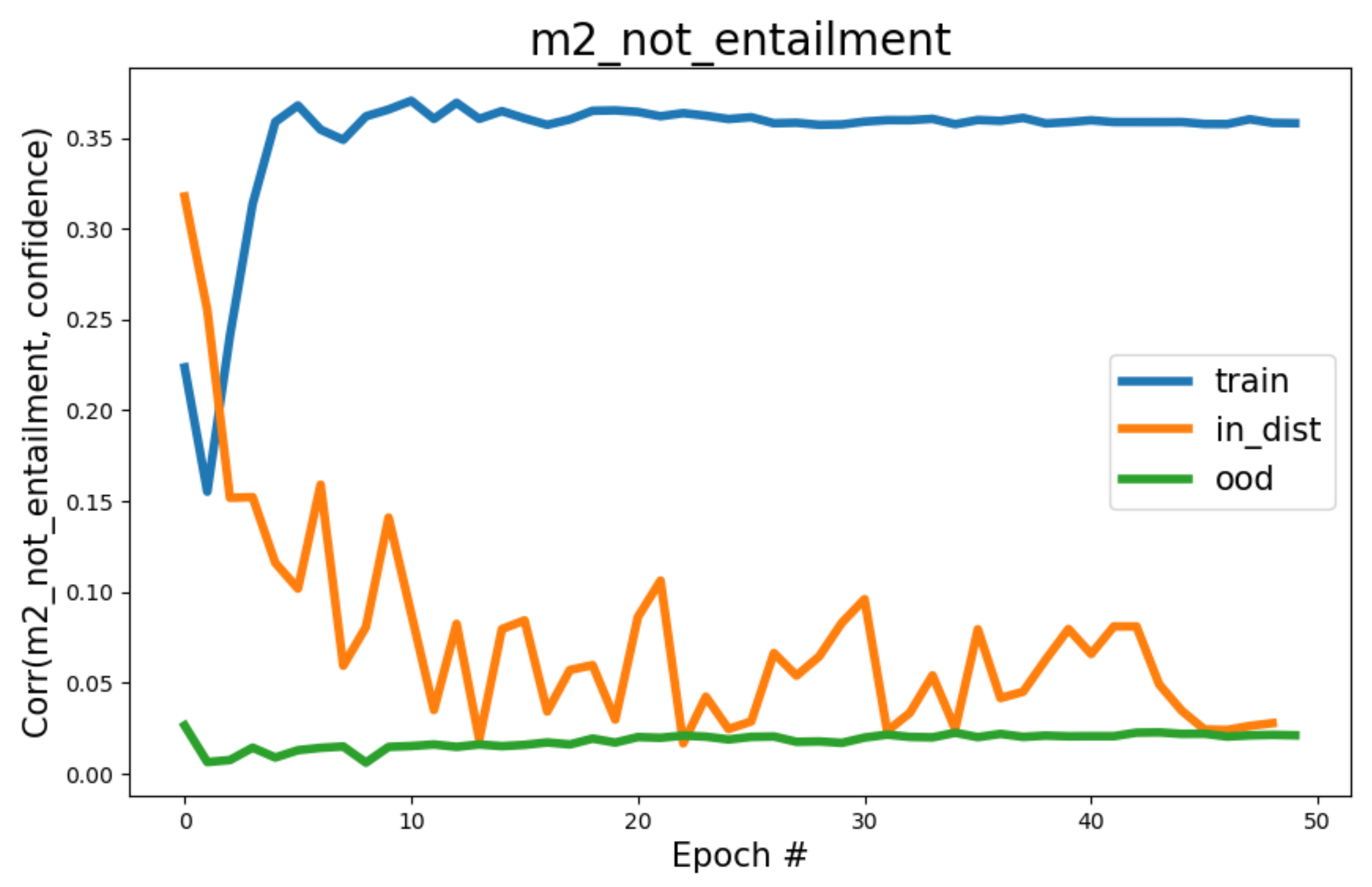}
    \caption{Train \& in-distribution: RTE, OOD: WNLI}
  \end{subfigure}
  \hfill
  \begin{subfigure}[t]{.32\textwidth}
    \centering
    \includegraphics[width=\linewidth]{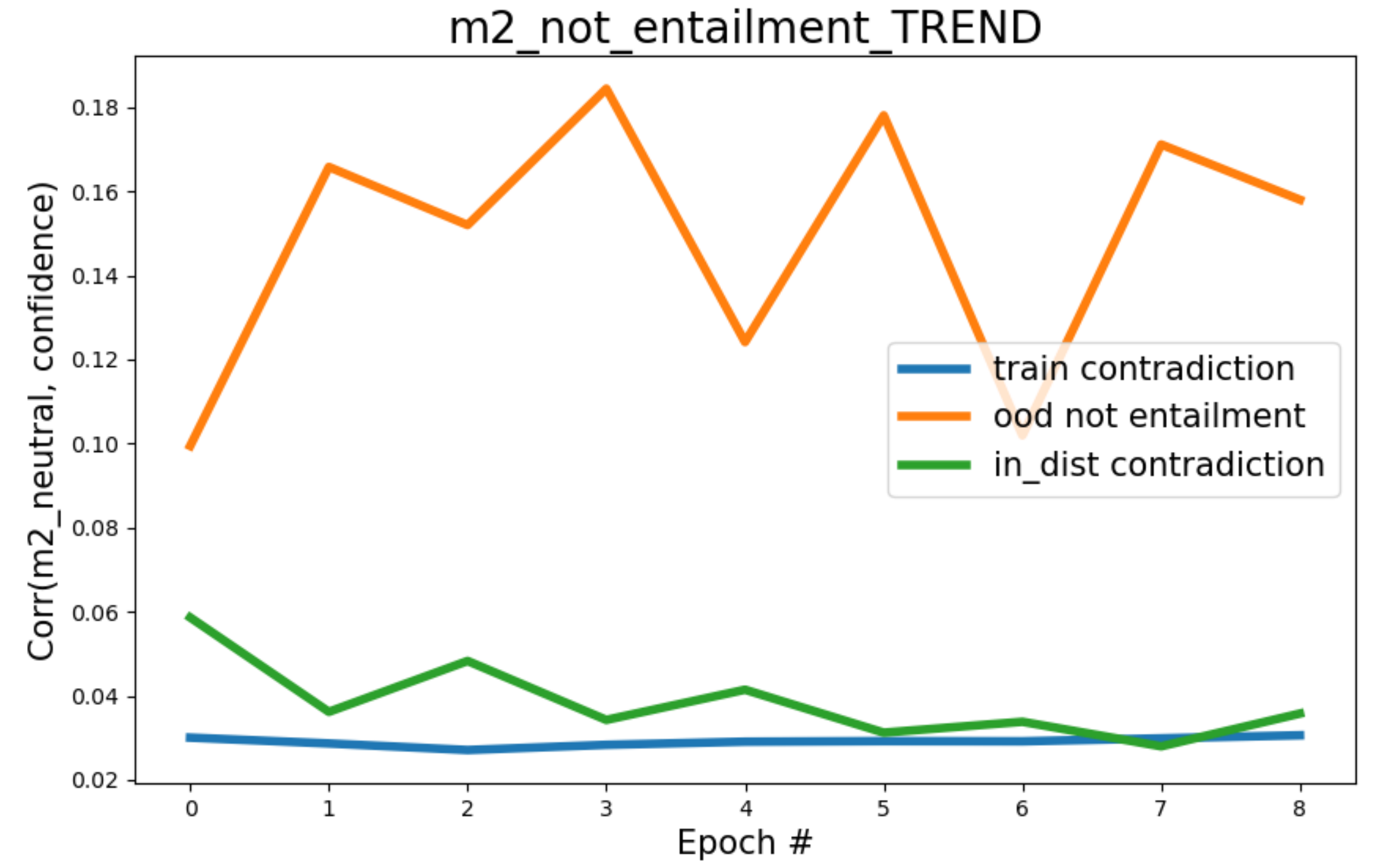}
    \caption{Train \& in-distribution: MNLI, OOD: RTE}
  \end{subfigure}
  \caption{Supplementary results for \ref{sec:result-ex2}. Correlation between $\hat{\mu}_i$ of \textbf{non-entailment} samples and m2}
\end{figure}

\subsection{Supplementary material: Extra lexical overlap measure}
\label{appendix:m1}
We also added another measure to quantify tendency to adopt lexical overlap heuristic. We calculated $m1 = \frac{\left | s1 \bigcap s2 \right |}{\left | s1 \right |}$. Essentially, this measures how much percentage of words found in the premise ($s1$) can also be found in the hypothesis ($s2$).
\begin{figure}[H]
  \begin{subfigure}[t]{.45\textwidth}
    \centering
    \includegraphics[width=\linewidth]{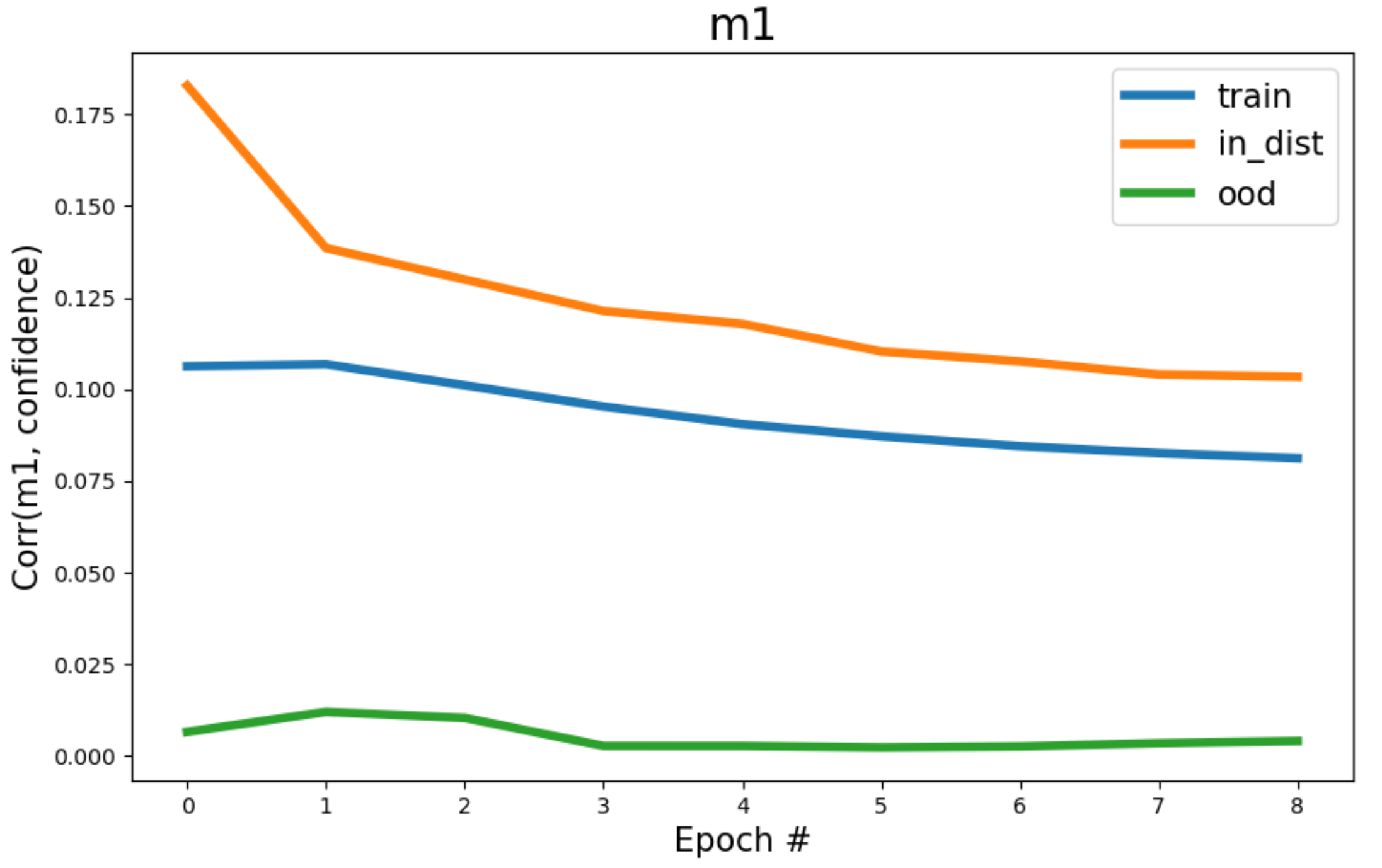}
    \caption{Correlation between \textbf{m2} and \textbf{all} samples $\hat{\mu}_i$}
  \end{subfigure}
  \hfill
  \begin{subfigure}[t]{.45\textwidth}
    \centering
    \includegraphics[width=\linewidth]{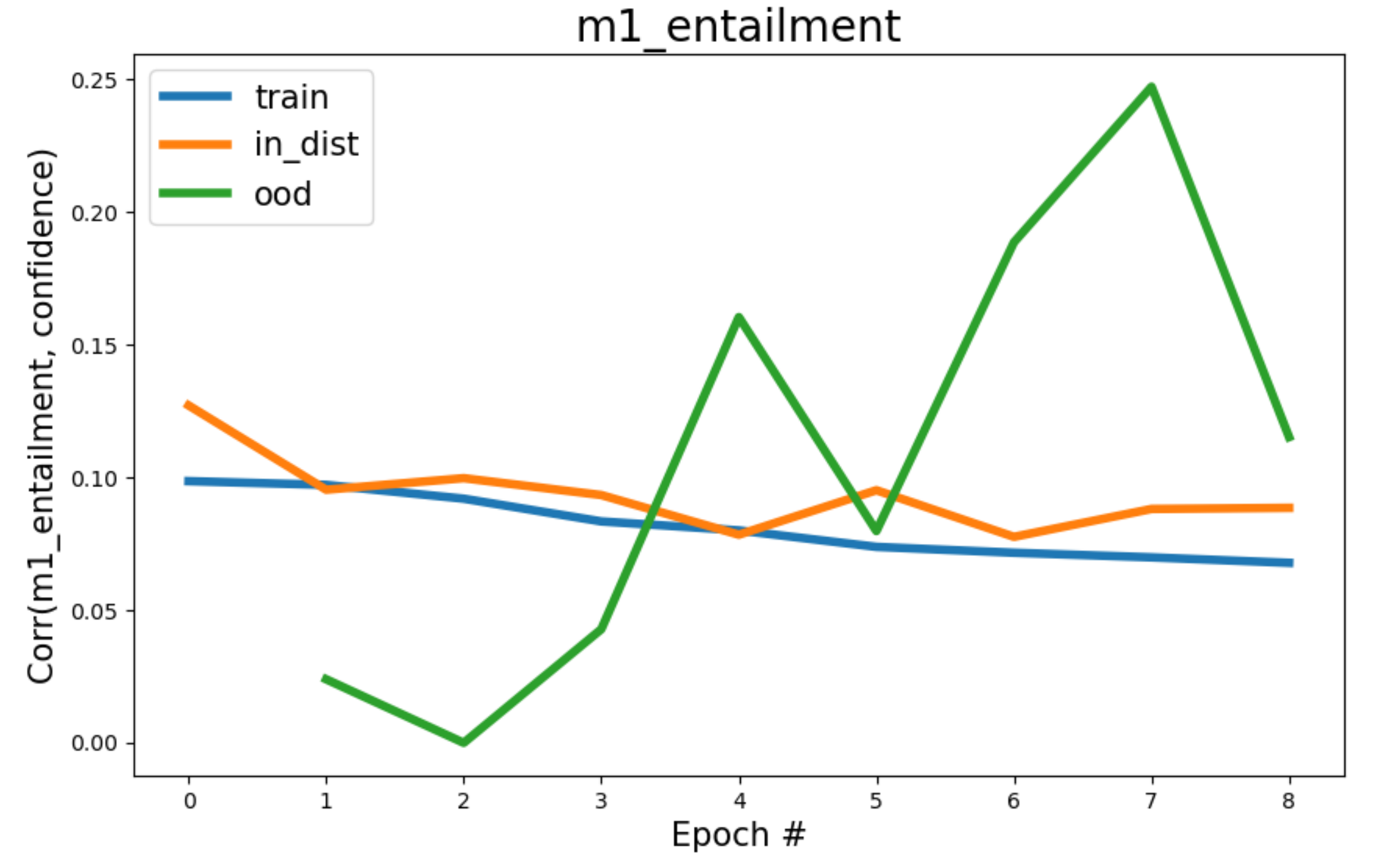}
    \caption{Correlation between \textbf{m2} and \textbf{entailment} samples $\hat{\mu}_i$}
  \end{subfigure}
  \caption{Results for hypothesis \ref{sec:hypothesis-heuristic}. Training and in-distribution test samples are MNLI, and OOD samples are WNLI.}
\end{figure}

\begin{figure}[H]
  \begin{subfigure}[t]{.45\textwidth}
    \centering
    \includegraphics[width=\linewidth]{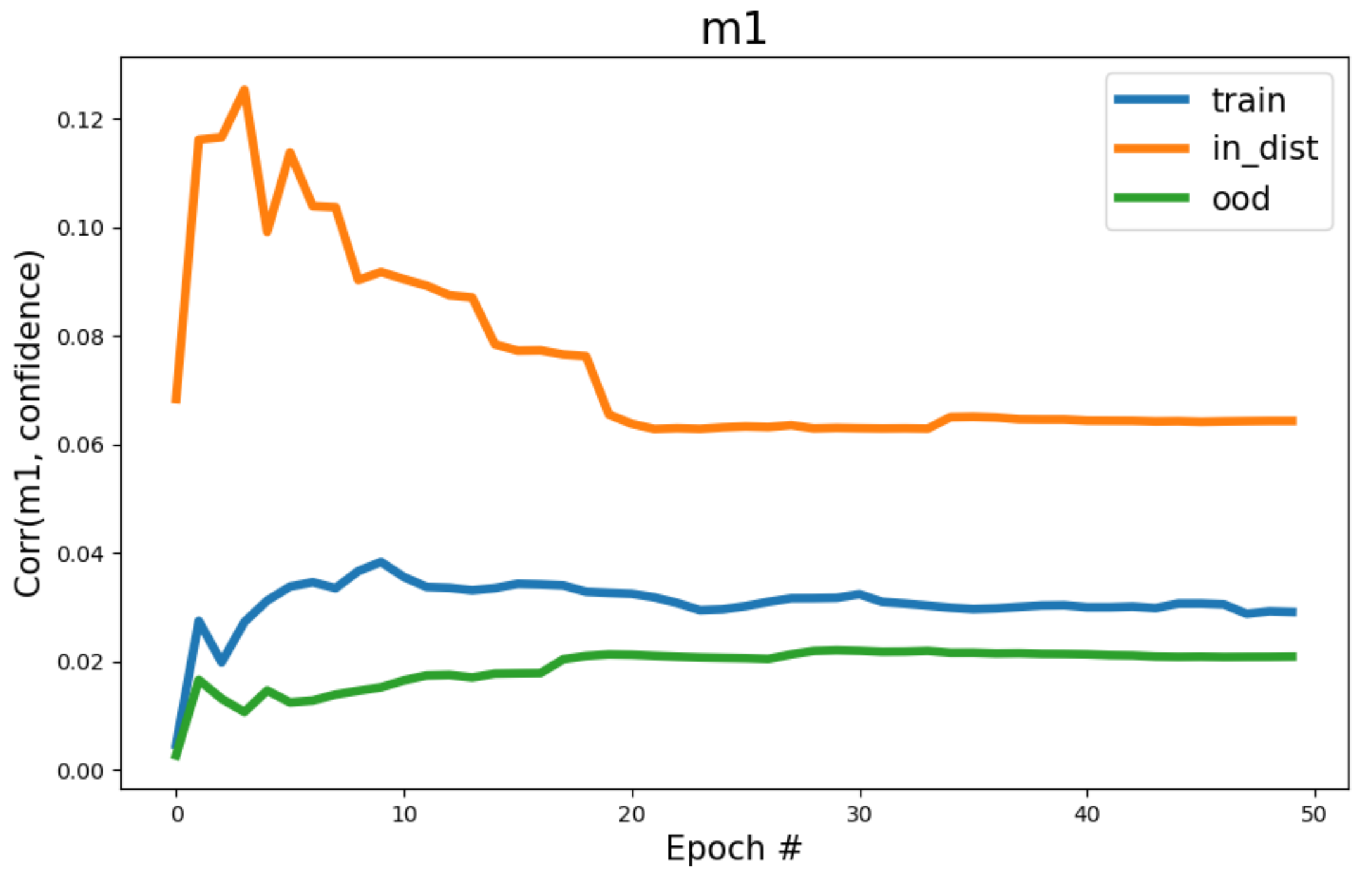}
    \caption{Correlation between \textbf{m2} and \textbf{all} samples $\hat{\mu}_i$}
  \end{subfigure}
  \hfill
  \begin{subfigure}[t]{.45\textwidth}
    \centering
    \includegraphics[width=\linewidth]{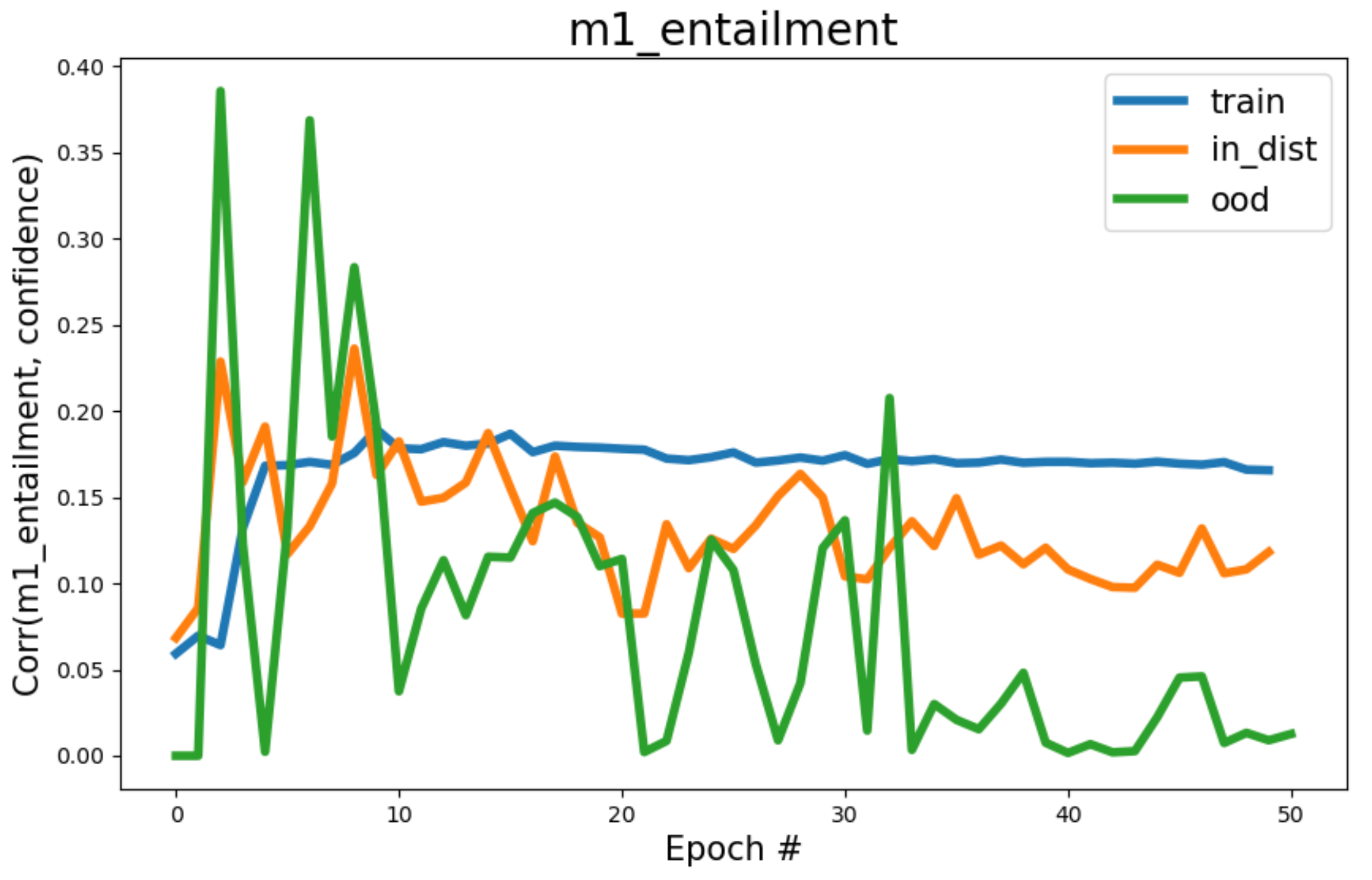}
    \caption{Correlation between \textbf{m2} and \textbf{entailment} samples $\hat{\mu}_i$}
  \end{subfigure}
  \caption{Results for hypothesis \ref{sec:hypothesis-heuristic}. Training and in-distribution test samples are RTE, and OOD samples are WNLI.}
\end{figure}

\begin{figure}[H]
  \begin{subfigure}[t]{.45\textwidth}
    \centering
    \includegraphics[width=\linewidth]{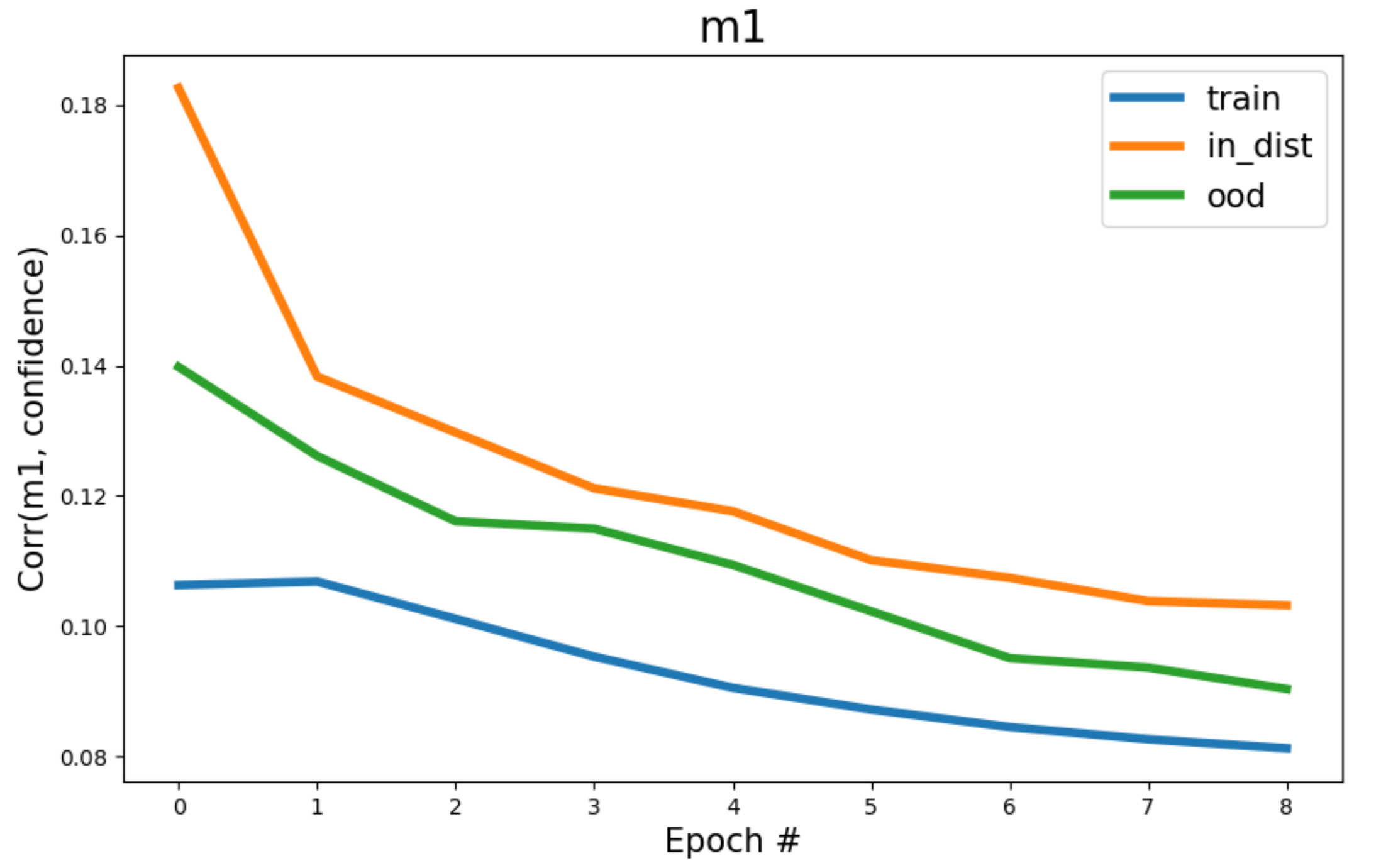}
    \caption{Correlation between \textbf{m2} and \textbf{all} samples $\hat{\mu}_i$}
  \end{subfigure}
  \hfill
  \begin{subfigure}[t]{.45\textwidth}
    \centering
    \includegraphics[width=\linewidth]{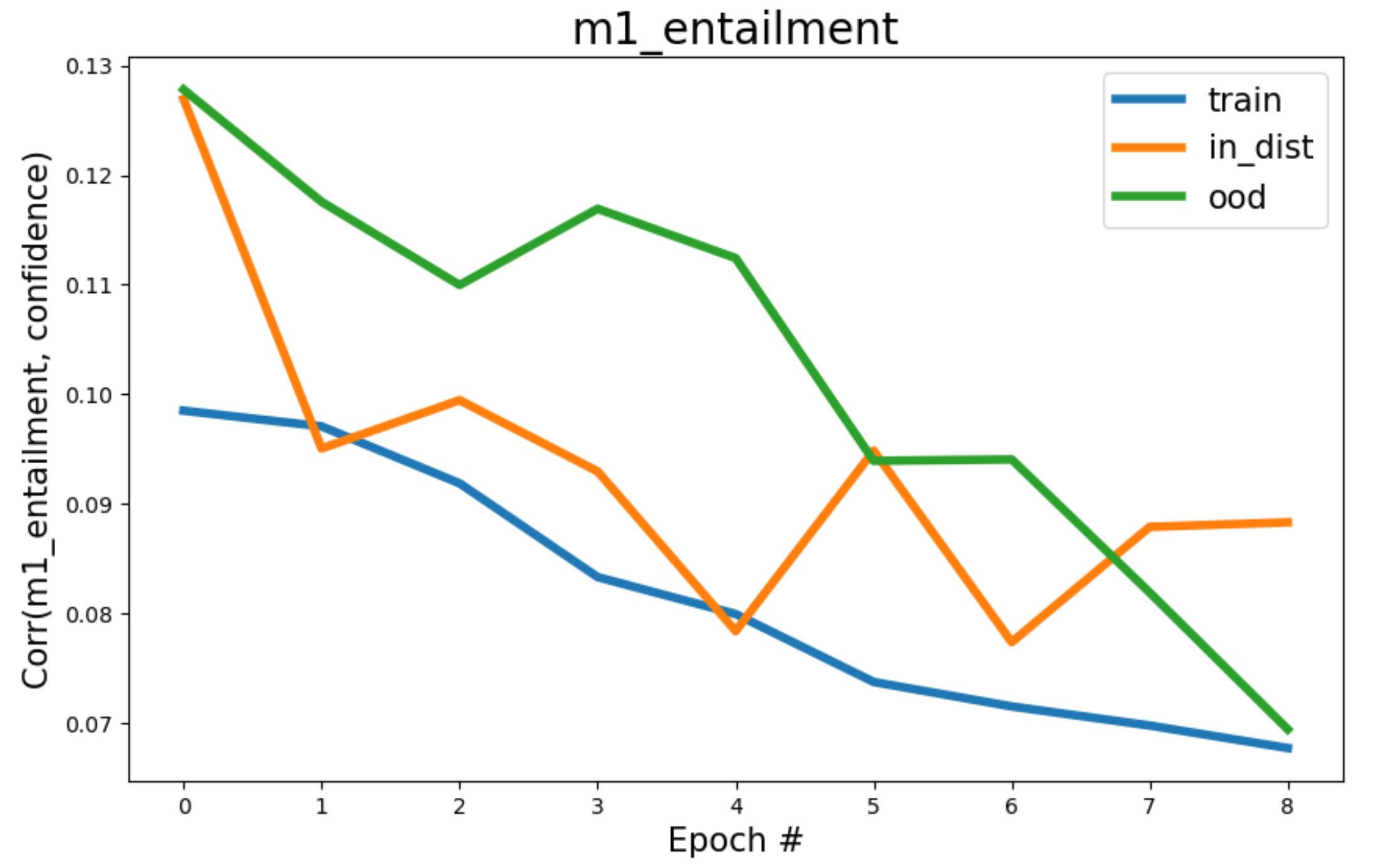}
    \caption{Correlation between \textbf{m2} and \textbf{entailment} samples $\hat{\mu}_i$}
  \end{subfigure}
  \caption{Results for hypothesis \ref{sec:hypothesis-heuristic}. Training and in-distribution test samples are MNLI, and OOD samples are RTE.}
\end{figure}

\end{document}